\documentclass[10pt,twocolumn,letterpaper]{article}

\usepackage{wacv}
\usepackage{times}
\usepackage{epsfig}
\usepackage{graphicx}
\usepackage{amsmath}
\usepackage{amssymb}

\usepackage{color}
\usepackage{epsfig}
\usepackage{epstopdf}
\usepackage{subfloat}
\usepackage{subcaption}
\usepackage{lineno}
\usepackage{float}
\usepackage{units}
\usepackage{bm}
\usepackage{enumerate}
\usepackage{enumitem}
\usepackage{multirow}
\usepackage{rotating}
\usepackage[hang,flushmargin]{footmisc}
\usepackage{mdwlist}
\usepackage{multirow}
\usepackage{lipsum}
\usepackage{textcomp}
\usepackage[table,xcdraw]{xcolor}
\usepackage{tablefootnote}
\usepackage{threeparttable}
\usepackage{algorithm}
\usepackage{algpseudocode}
\usepackage[table,xcdraw]{xcolor}

\usepackage[T1]{fontenc}
\usepackage[utf8]{inputenc}
\usepackage[noblocks]{authblk}

\usepackage[pagebackref=true,breaklinks=true,letterpaper=true,colorlinks=true,bookmarks=false,bookmarksdepth=3]{hyperref}

\hypersetup{
	colorlinks=true,       
	linkcolor=blue,          
	citecolor=red,        
	urlcolor=magenta           
}

\wacvfinalcopy

\def\wacvPaperID{1263} 

\ifwacvfinal\fi
\begin{document}
	

\title{\vspace{-0.8cm}
	Animal Detection in Man-made Environments
}

\author[1]{Abhineet Singh
}
\author[2]{Marcin Pietrasik
}
\author[2]{Gabriell Natha
}
\author[2]{Nehla Ghouaiel
}
\author[2]{Ken Brizel
}
\author[1]{Nilanjan Ray
}
%
\affil[1]{Department of Computing Science, University of Alberta}
\affil[2]{Alberta Centre for Advanced MNT Products (ACAMP)}

\maketitle
\ifwacvfinal\thispagestyle{empty}\fi

\begin{abstract}
	Automatic detection of animals that have strayed into human inhabited areas has important security and road safety applications.
	This paper
	attempts to solve
	this problem using
	deep learning techniques from a variety of computer vision fields including object detection, segmentation, tracking and edge detection.
	Several interesting insights into transfer learning are elicited while adapting models trained on benchmark datasets for real world deployment.
	Empirical evidence is presented to demonstrate the inability of detectors to generalize from training images of animals in their natural habitats to deployment scenarios of man-made environments.
	A solution is also proposed using semi-automated synthetic data generation for domain specific training.
	Code and data used in the experiments are made available to facilitate further work in this domain.

\end{abstract}

\section{Introduction}
\label{sec_intro}

Object detection is an important field in computer vision that has seen very rapid improvements in recent years using deep learning \cite{Szegedy2013DeepNN,Huang2016SpeedAccuracyTF,Liu18_review}.
Most detectors are trained and tested on benchmark datasets like COCO \cite{TsungYi14_coco}, Open Images \cite{Kuznetsova18ax_OpenImages}, KITTI \cite{Geiger2013IJRR_kitti} and VOC \cite{Everingham10_voc}.
In order to apply these in a particular domain like animal detection, a model pre-trained on one of these datasets is fine-tuned on domain-specific data,
usually
by training
only the last few layers.
This is known as transfer learning \cite{Pan2010ASO_transfer, Yosinski14_nips_transfer} and is often enough to obtain
good
performance in the new domain
as long as it does not differ drastically from the original.
The goal of this work is to use transfer learning to adapt state of the art object detection methods for detecting several types of large Alberta animals in real-time video sequences captured from one or more monocular cameras
in moving ground vehicles.
The animals that most commonly stray into human habitations include: deer, moose, coyotes, bears, elks, bison, cows and horses.
There are two deployment scenarios:
\begin{itemize}[leftmargin=*]
	\itemsep0em 
	\item Detecting threats in an autonomous all-terrain vehicle (ATV) patrolling the Edmonton International Airport perimeter for security and surveillance purposes.
	\item Finding approaching animals in side–mounted cameras on buses plying the Alberta highways to issue a timely warning to the driver for collision avoidance.
\end{itemize}

The main challenge here is the scarcity of existing labeled data with sufficient specificity
to the target domain to yield good models by fine-tuning pre-trained detection networks.
Although several of the large public datasets like COCO \cite{TsungYi14_coco} do include some of the more common animals like bears and horses, these rarely include the Canadian varieties that are the focus of this work and often feature incorrect backgrounds.
Even larger classification datasets like Imagenet \cite{imagenet_cvpr09} do include images of many of the target animals but only provide image level labels so the bounding boxes would have to be added manually.
There are also several animal specific datasets \cite{lilabc19_ecology_dataset,gray19_deep_ecology} but these likewise do not match the target requirements well, having, for example, aerial viewpoints \cite{Kellenberger2018_dataset_uav,Kellenberger18_dataset_uav}, incorrect species \cite{Khosla_FGVC2011_dataset_dog,parkhi12a_iiit_pet_dataset,Li19_tiger_dataset,WelinderEtal2010_caltech_usd_birds,Horn19_nabirds_dataset,Swanson2015_serengeti_dataset,Parham18_wild_dataset} or habitats \cite{Beery2018_caltech_trap_dataset,beery2019iwildcam} and no bounding box annotations \cite{Horn18_inaturalist,Swanson2015_serengeti_dataset,Tabak18_camera_trap_north_america,Willi18_camera_trap_data}.

The lack of training data
was addressed by collecting and labelling a sufficiently large number of images of the target animals.
This was initially confined to videos
since labeling these was easier to semi-automate (Sec. \ref{sec_data_collection}) and training detectors on videos showing the animals in a variety of poses seemed to concur better with deployment on camera videos captured from moving vehicles.
However, tests showed that detection performance is far more sensitive to the range of \textit{backgrounds} present in the training set rather than variations in the appearance of the animal itself (Sec. \ref{sec_results}).
Though static images helped to resolve this to a certain extent, they are much harder to obtain in large numbers and a lot more time-consuming to label.
More importantly, neither static nor video images of animals are easy to acquire with the kinds of structured man-made surroundings that the airport perimeter and highways present.
This paper thus proposes a solution based on synthetic data generation using a combination of interactive mask labelling, instance segmentation and automatic mask generation (Sec. \ref{sec_synthetic}).

Another significant challenge is the need for the detector to be fast enough to process streams from up to 4 cameras in real time while
running on relatively low-power machines since both deployment scenarios involve mobile computation where limited power availability makes it impractical to run a multi-GPU system.
This is addressed using RetinaNet \cite{Lin2017FocalLF} and
YOLOv3 \cite{Redmon18_yolov3} which turned out to be surprisingly competitive with respect to much slower models.
To the best of our knowledge, this is also the first large-scale study of applying deep learning for animal detection in general and their Canadian varieties in particular.
It presents interesting insights about transfer learning gained by training and testing the models on static, video and synthetic images in a large variety of configurations.
Finally, it provides practical tips that might be useful for real world deployment of deep learning models.
Code and data are made publicly available to facilitate further work in this field \cite{animal_detection_github}.

\section{Related Work}
\label{sec_related_work}

Animal recognition in natural images is a well researched area with applications mostly in ecological conservation.
As in the case of available data, most of the existing work is not closely allied to the domain investigated in this paper.
Three main categories of methods can be distinguished from the literature corresponding to the type of input images used.
The first category corresponds to aerial images captured from unmanned aerial vehicles (UAVs).
A recent work \cite{Kellenberger19_uav} introduced an active learning \cite{active_learning11} method called transfer sampling that uses optimal transport \cite{opt_transport13} to handle domain shift between training and testing images that occurs when using training data from previous years for target re-acquisition in follow-up years.
This scenario is somewhat similar to the current work so transfer sampling might have been useful here but most of this work had already been done by the time \cite{Kellenberger19_uav} became available.
Further, it would need to be reimplemented since its code is not released and the considerable domain difference between aerial and ground imagery is likely to make adaptation difficult.
Finally, most domain adaptation methods, including \cite{Kellenberger19_uav}, require unlabeled samples from the target domain which are not available in the current case.
Other examples of animal detection in UAV images include \cite{Kellenberger17_uav,Rey17_uav,Kellenberger18_dataset_uav,Kellenberger2018_dataset_uav} but, like \cite{Kellenberger19_uav}, all of these are focused on African animals.

The second category
corresponds to motion triggered camera trap images.
These have been reviewed in \cite{Schneider19_trap_review} and \cite{Sara18_trap} where the latter reported similar difficulties in generalizing to new environments as were found here.
The earliest work using deep learning was \cite{Chen14_trap} where graph cut based video segmentation is first used to extract the animal as a moving foreground object and then a classification network is run on the extracted patch.
A more recent work \cite{Schneider18_trap} that most closely resembles ours tested two detectors - Faster RCNN \cite{Shaoqing15_frcnn} and YOLOv2 \cite{Redmon2016_yolov2} - 
and reported respective accuracies of $ 93\% $ and $ 76\% $.
However,
the evaluation criterion used there is more like classification than detection since it involves computing the overlaps of all detected boxes with the ground truth and then comparing the class of only the maximum overlap detection to decide if it is correct.
Other recent works in this category, most of them likewise dealing mainly with classification, include \cite{Norouzzadeh2018_trap,Willi18_camera_trap_data,Tabak18_camera_trap_north_america,Hayder18_trap,Yousif17_trap,Zhu17_trap}.

The third category,
which includes this work, involves real-time videos captured using ground-level cameras.
An important application of such methods is in ethology for which many general purpose end-to-end graphical interface systems have been developed \cite{Mnck2018BioTrackerAO,Sridhar19_vid_Tracktor,Werkhoven19_vid_margo,Ferrero19_vid_idtracker,Patman18_BioSense}.
Methods specialized for particular species like cows \cite{Zin16_vid_cow}, beef cattle \cite{TerSarkisov2018_vid_beef_cattle} and tigers \cite{Li19_tiger_dataset} have also been proposed where the latter includes re-identification that is typically done using camera trap images.
Surveillance and road safety applications like ours are much rarer in the literature and it seems more common to employ non-vision sensors and fencing/barrier based solutions, probably because many animal vehicle collisions happen in the dark \cite{Wilkins2019_avc}. 
Examples include infrared images \cite{Forslund14_vid_night}, thermal and motion sensors \cite{Grace2017_avc}, ultra wide band wireless sensor network \cite{Xue17_avc_wireless} and kinect \cite{Zhu15_kinect}.

A general review of the vision techniques used in this work including object detection, segmentation and tracking are excluded here due to space constraints.
The actual methods used in the experiments are detailed in Sec. \ref{sec_methodolgy}.


\begin{figure*}[!htbp]
	\includegraphics[width=0.246\textwidth]{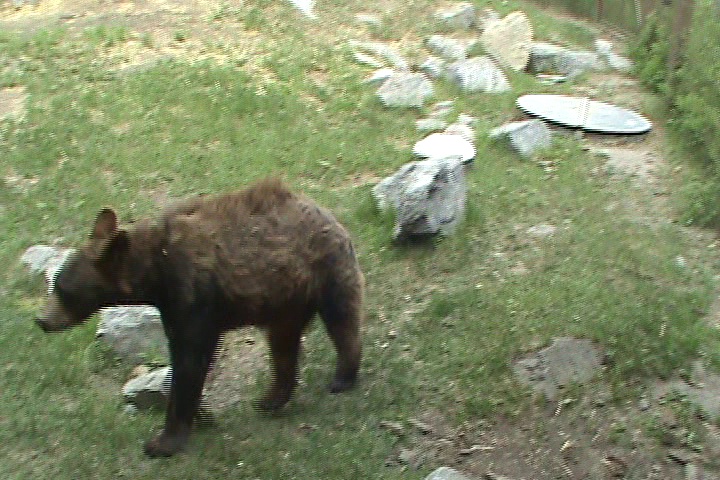}
	\includegraphics[width=0.246\textwidth]{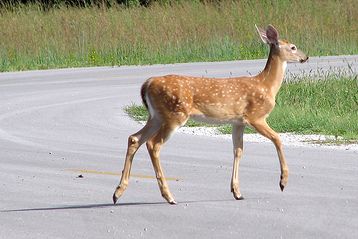}
	\includegraphics[width=0.246\textwidth]{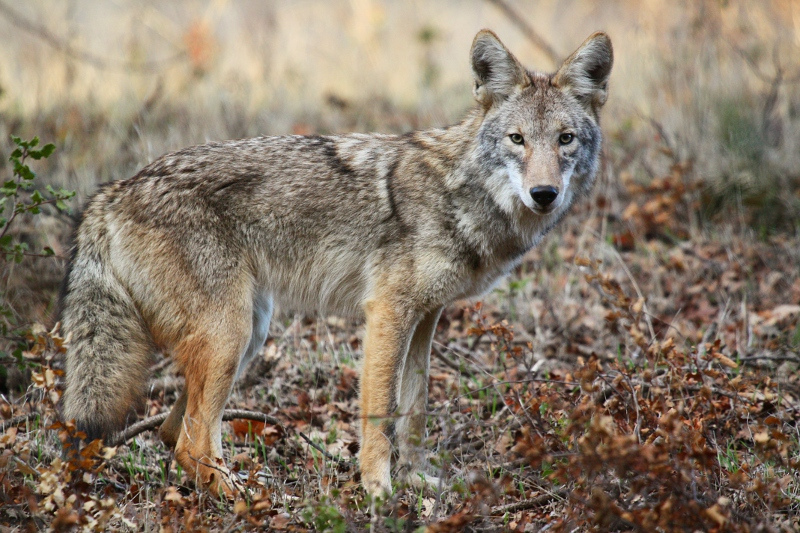}
	\includegraphics[width=0.246\textwidth]{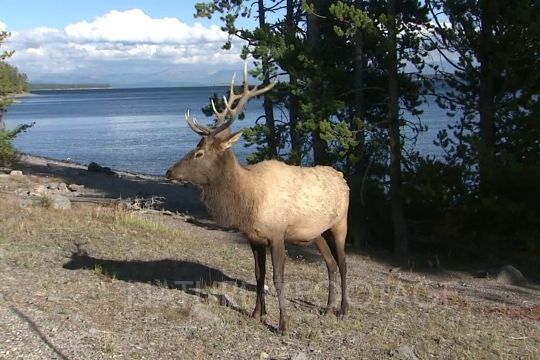}
	\includegraphics[width=0.246\textwidth]{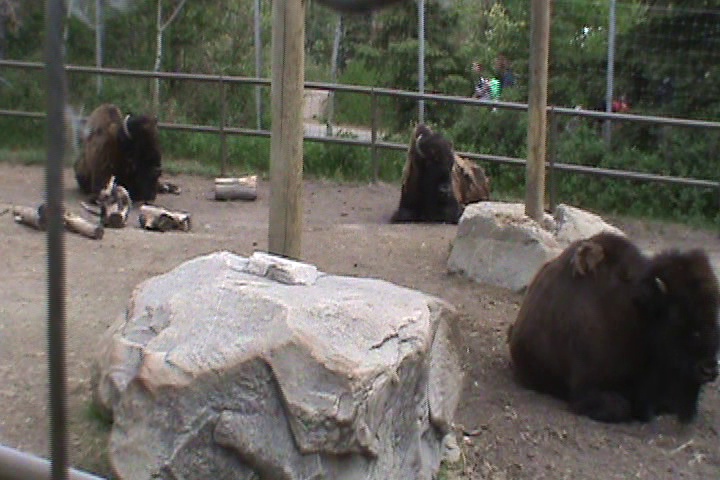}
	\includegraphics[width=0.246\textwidth]{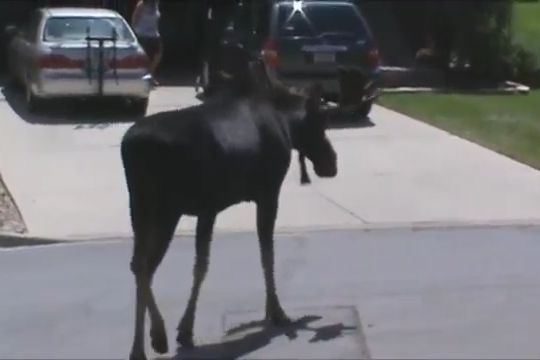}
	\includegraphics[width=0.246\textwidth]{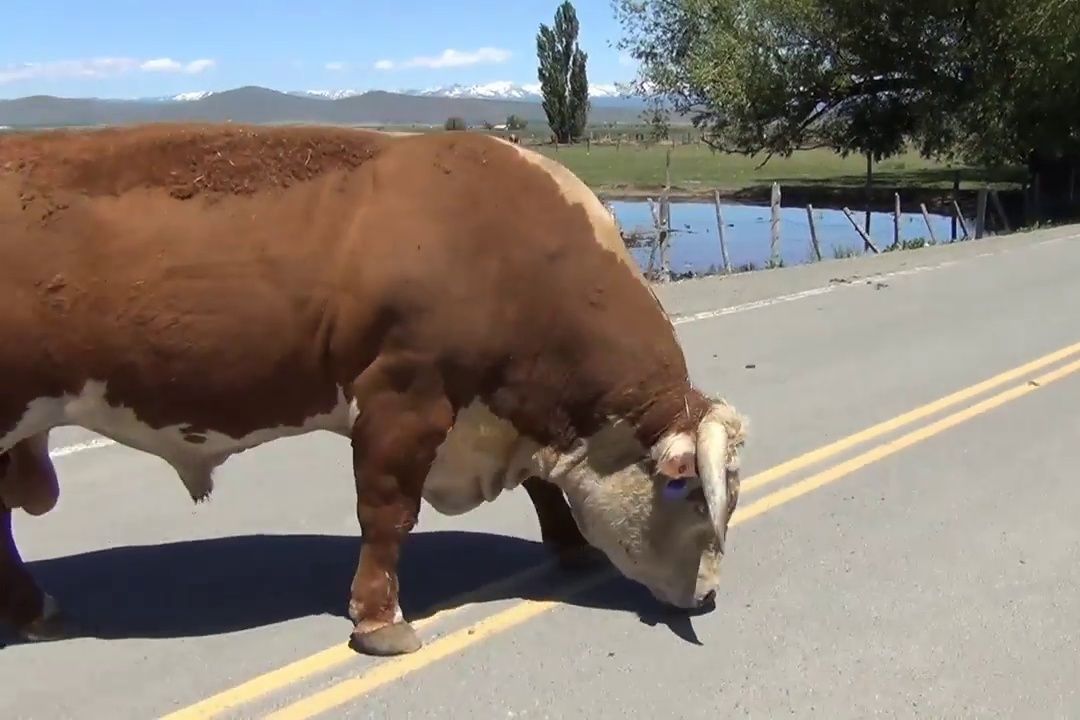}
	\includegraphics[width=0.246\textwidth]{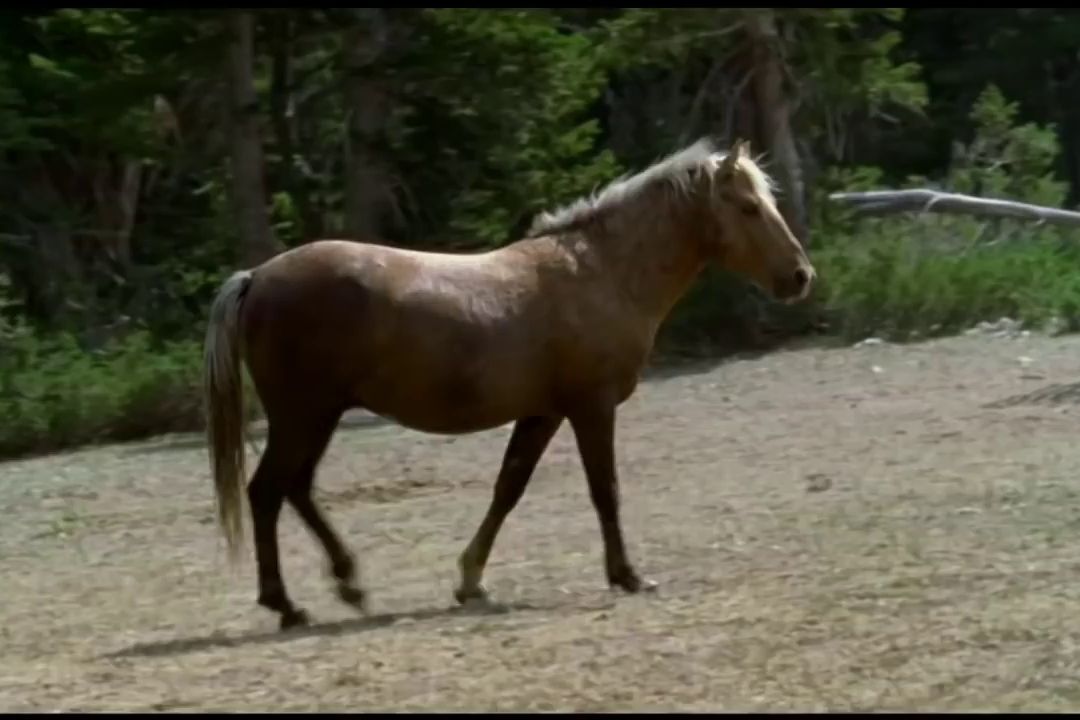}
	\caption{Sample collected images: (clockwise from top left) bear (Calgary Zoo), deer (Google Images), coyote (Google Images), elk (Nature Footage),  horse (YouTube), cow (YouTube), moose (YouTube) and bison (Calgary Zoo)}
	\label{fig_footage}
\end{figure*}

\begin{table}[]
	\centering
	\caption{Annotation counts
		(\textbf{seq}, \textbf{syn}: sequences, synthetic)
	}
	{
		\footnotesize
	\begin{tabular}{|l|l|l|l|l|l|l|}
		\hline
		\multirow{2}{*}{\textbf{Class}} & \multicolumn{2}{c|}{\textbf{Videos (Real)}} & \multicolumn{3}{c|}{\textbf{Static Images}}         & \multirow{2}{*}{\textbf{Total}} \\ \cline{2-6}
		& \textbf{Seq}     & \textbf{Images}    & \textbf{Real} & \textbf{Syn} & \textbf{Total} &                                 \\ \hline
		\textbf{Bear}                   & 92                     & 25715              & 1115          & 286                & 1401           & 27116                           \\ \hline
		\textbf{Bison}                  & 88                     & 25133              & 0             & 0                  & 0              & 25133                           \\ \hline
		\textbf{Cow}                    & 14                     & 5221               & 0             & 0                  & 0              & 5221                            \\ \hline
		\textbf{Coyote}                 & 113                    & 23334              & 1736          & 260                & 1996           & 25330                           \\ \hline
		\textbf{Deer}                   & 67                     & 23985              & 1549          & 286                & 1835           & 25820                           \\ \hline
		\textbf{Elk}                    & 78                     & 25059              & 0             & 0                  & 0              & 25059                           \\ \hline
		\textbf{Horse}                  & 23                     & 4871               & 0             & 0                  & 0              & 4871                            \\ \hline
		\textbf{Moose}                  & 97                     & 24800              & 0             & 260                & 260            & 25060                           \\ \hline
		\textcolor[rgb]{ 0,  .439,  .753}{\textbf{Total}}                  & \textcolor[rgb]{ 0,  .439,  .753}{\textbf{572}}           & \textcolor[rgb]{ 0,  .439,  .753}{\textbf{158118}}    & \textcolor[rgb]{ 0,  .439,  .753}{\textbf{4400}} & \textcolor[rgb]{ 0,  .439,  .753}{\textbf{1092}}      & \textcolor[rgb]{ 0,  .439,  .753}{\textbf{5492}}  & \textcolor[rgb]{ 0,  .439,  .753}{\textbf{163610}}                 \\ \hline
	\end{tabular}
	}
\label{tab_annotations}
\end{table}

\section{Methodology}
\label{sec_methodolgy}

\subsection{Data Collection}
\label{sec_data_collection}

To facilitate the large number of training images needed, a combination of video and static images was used. 
Video was collected both directly
with handheld video cameras around Calgary area, such as the Calgary Zoo, as well as online via YouTube and Nature Footage \cite{nature_footage}.
Due to the large quantity of static images that was required, downloading them one by one was not feasible.
Instead, ImageNet \cite{imagenet_cvpr09}
was used
as it provides a searchable database of images with links whereby they can be downloaded in bulk using scripts.
However, not all animal species are available there and not all available ones have enough images.
Google Images was thus also used by searching for specific taxonomic classification and downloading the results in bulk using browser extensions.
After downloading static images, it was necessary to
verify that all were of the intended animal and remove any mislabeled or unhelpful images.
Figure \ref{fig_footage} shows sample images of all animals while Table \ref{tab_annotations} provides quantitative details.

\begin{figure*}[t]

	\includegraphics[width=\textwidth]{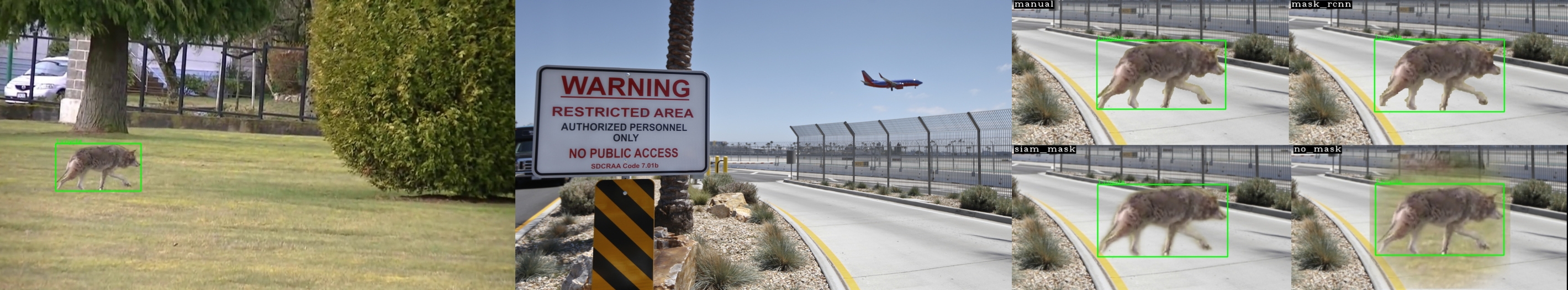}
	\includegraphics[width=\textwidth]{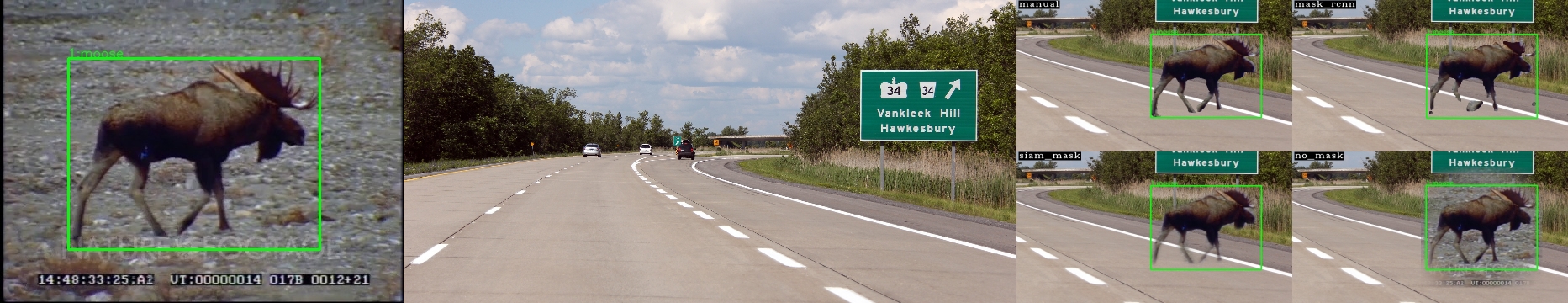}

	\caption{
		Synthetic data samples with corresponding source and target images for (top) coyote on airport and (bottom) moose on highway.
		Each row shows (left to right) animal source image, target background image and crops of synthetic images generated using (clockwise from top left) manual labeling, Mask RCNN, SiamMask and Gaussian blending (no mask).
	}
	\label{fig_syn}
\end{figure*}

\subsection{Labeling}
\label{sec_labeling}

\subsubsection{Bounding Boxes}
\label{sec_boxes}

Annotation was done using a heavily modified version of an open source image annotation tool called LabelImg \cite{Tzutalin15_LabelImg}.
This tool takes a video file or sequence of images as input and allows the users to annotate bounding boxes with class labels over them.
The SiamFC tracker \cite{Luca16_siamfc} was integrated with the tool to make video annotation semi–automated so that the user only needs to manually annotate the animal in the first frame, track it till the tracker starts drifting, fix the box in the last tracked frame, start tracking again and repeat this process till all frames are labeled.

\subsubsection{Segmentation Masks}
\label{sec_masks}

Pixel wise masks were needed to generate high-quality synthetic data (Sec. \ref{sec_synthetic}).
Annotation tools that support masks do exist \cite{dutta2019vgg,dutta2016via,labelme2016,Breheret17_annotation,Russell19_LabelMe,RectLabel,Labelbox}, including AI assisted services \cite{playment}, but all have issues such as too course masks \cite{dutta2019vgg,dutta2016via,labelme2016,Breheret17_annotation}, Linux incompatibility \cite{RectLabel}, paid or propriety license \cite{playment,Labelbox} or cloud-data restriction \cite{Russell19_LabelMe}.
Also, it was desirable to semi-automate mask generation using the already existing bounding boxes
which is not allowed by any of the tools.
Mask annotation functionality was thus added to the labelling tool
with support for 3 different modalities to add or refine masks - drawing, clicking to add boundary points and painting with variable sized brushes.


Semi-automated mask generation was done using a combination of motion based interpolation, edge detection and tracking.
An approximate mask is generated for a given frame by estimating the motion between its bounding box and that in a previous frame whose mask has already been annotated.
In addition, holistically nested edge detection (HED) \cite{Xie15_hed} followed by adaptive thresholding is used to obtain a rough boundary of the animal that can be refined by painting.
Finally, the SiamMask tracker \cite{wang2019_siam_mask}, that outputs both bounding boxes and segmentation masks, was integrated with the labelling tool to generate low-quality masks in a fully automated manner.
Mask labelling was a slow and laborious task and took anywhere from 1 - 8 minutes per frame depending on animal shape and background clutter.
An arguably more sophisticated pipeline for rapid generation of segmentation masks has recently been proposed \cite{Benenson2019LargescaleIO}.
However, it became available too late to be utilized in this project, does not provide a publicly available implementation and its proposed pipeline includes human involvement on too large a scale to be practicable here.
A recent video mask prediction method \cite{semantic_cvpr19} likewise came out too late and also rendered unnecessary by
SiamMask.

\subsection{Object Detection}
\label{sec_detection}

Object detection has improved considerably since the advent of deep learning \cite{Szegedy2013DeepNN} within which two main categories of detectors have been developed.
The first category includes methods based on the RCNN architecture \cite{Girshick14_rcnn,Girshick2016_rcnn} that utilize a two-step approach.
A region proposal method is first used to generate a large number of class agnostic bounding boxes that show high probability of containing a foreground object.
Some of these are then processed by a classification and regression network to give the final detections.
Examples include Fast \cite{Girshick15_fast_rcnn} and Faster \cite{Shaoqing15_frcnn,Ren17_frcnn} RCNN and RFCN \cite{Dai16_rfcn}.
The second category includes methods that combine the two steps into a single end to end trainable network.
Examples include YOLO \cite{Redmon15_yolo,Redmon2016_yolov2,Redmon18_yolov3}, SSD \cite{Liu16_ssd} and RetinaNet \cite{Lin2016FeaturePN,Lin2017FocalLF}.
Apart from its high-level architecture, the performance of a detector also depends
on the backbone network used for feature extraction.
Three of the most widely used families of performance-oriented backbones include ResNet \cite{He16_resnet,He16_resnet2,He16_resnet1k},  Inception \cite{Szegedy15_inception,Szegedy16_inceptionv2,Chollet17_xception,Szegedy17_inceptionv4} and Neural Architecture Search (NAS) \cite{Zoph17_nas_rl,Zoph18_nas}.
Several architectures have also been developed with focus on high speed and low computational requirements.
The most widely used among these are the several variants of MobileNet \cite{Weyand17_mobilenet,Sandler18_mobilenetv2,Howard19_mobilenetv3}.

Five high level detector architectures have been used here – Faster RCNN, RFCN, SSD, RetinaNet and YOLO.
Three different backbone networks are used for Faster RCNN - ResNet101, InceptionResnetv2, NAS - and two for SSD - Inceptionv2 \cite{Szegedy16_inceptionv2}, Mobilenetv2 \cite{Sandler18_mobilenetv2} - for a total of 8 detectors. 
ResNet101 and ResNet50 are used as backbones for RFCN and RetinaNet respectively.
All 3 variants of YOLO \cite{Redmon15_yolo,Redmon2016_yolov2,Redmon18_yolov3} were experimented with, though only YOLOv3 \cite{Redmon18_yolov3} results are included here as being the best performer. 
These methods were chosen to cover a good range of accuracies and speeds among modern detectors.

All of the above are \textit{static} detectors that process each frame individually without utilizing the temporal correlation inherent in video frames.
Detectors have also been developed to incorporate this information for reducing missed detections due to issues like partial occlusions and motion blur.
Examples include Seq-NMS \cite{Han16_seqnms}, TCNN \cite{Kang16_tcnn,Kang18_tcnn}, TPN \cite{Kang17_tpn}, D\&T \cite{Feichtenhofer17_dt} and FGFA \cite{Zhu17_fgfa,Zhu17_deep_feature_flow}.
However, none of these have compatible implementations and most need either optical flow, patch tracking or both to run in parallel with a static detector which makes them too slow to be used here.
LSTM-SSD \cite{Liu18_vid_lstm,Liu19_vid_lstm} is the only recent video detector that is both fast and open source 
but
attempts to incorporate this here showed its implementation \cite{lstm_ssd19} to be too buggy and poorly documented to be usable without significant reimplementation effort not warranted by the modest improvement it seemed likely to provide.
Instead, a simple algorithm was devised to combine the
DASiamRPN
tracker \cite{Zhu18_dasiamrpn} with YOLO (Sec. \ref{sec_res_tracking})
to
gauge the potential benefit of temporal information
in videos.

\subsection{Synthetic Data Generation}
\label{sec_synthetic}

Experiments showed that detectors have limited ability to generalize to new backgrounds (Sec. \ref{sec_res_generalize}).
A solution considered first was to collect static images with as much background variation as possible to cover all target scenarios.
This proved to be impracticable due the difficulty of finding and labeling sufficient quantities of static images, exacerbated by our target scenarios consisting of man-made environments where it is extremely rare to find animals at all.
As a result, synthetic data was generated by extracting animals from existing labeled images
and adding them to images of the target backgrounds.
Attempts were initially made to do this without masks
by selecting
only the best matching source images for each target background
through techniques like histogram matching
and then
using
Gaussian blending
to smoothen the transition from source to target background.
However, this
failed to generate images that could either be perceived as realistic by humans or improve detection performance
(Sec. \ref{sec_res_synthetic}).
Pixel wise masks were therefore generated by manually labelling a sparse collection of frames with as much background variation as possible and then training instance segmentation models (Sec. \ref{sec_segmentation}) to automatically generate masks for remaining frames with similar backgrounds.
SiamMask tracker \cite{wang2019_siam_mask} was also used towards the end of the project to make
this
process
fully automated.
Generating synthetic images was much faster than labelling masks and only took about 1-10 seconds/frame.
Most of the effort was focused on generating static images since experiments (Sec. \ref{sec_res_vid_worth}) showed that videos do not help to improve detectors much.
It is also significantly harder to generate realistic videos as that requires camera motion in the source and target video clips to be identical.
Images were generated from 14 airport and 12 highway backgrounds with 11 source images for bears and deer, and 10 for coyotes and moose.
Fig. \ref{fig_syn} shows examples.

\subsection{Instance Segmentation}
\label{sec_segmentation}
Instance segmentation distinguishes between each instance of an object
as opposed to semantic segmentation that only identifies categories of objects.
The former intuitively seems more suitable for extracting animal boundaries from bounding boxes since it uses object level reasoning whereas the latter is more oriented towards pixel-level
classification.
This was confirmed by experiments with several state of the art semantic segmentation methods, including DeepLab \cite{ChenPKMY18_deeplab,deeplabv3plus2018}, UNet \cite{Ronneberger15_unet} and SegNet \cite{BadrinarayananK17_segnet}.
All of these generated masks that were too fine-grained to  cleanly segment out the animal from its background, instead producing many small groups of background pixels inside the animal and, conversely, animal pixels within the background.
Three instance segmentation methods were then considered – SharpMask/DeepMask \cite{Pinheiro16_sharpmask,Pinheiro15_deepmask}, Mask RCNN \cite{He17_mask_rcnn} and FCIS \cite{Li17_fcis}.
Mask RCNN was found to
produce the highest quality masks
so only its results are included.

\begin{figure*}[!htbp]
	\includegraphics[width=0.33\textwidth]{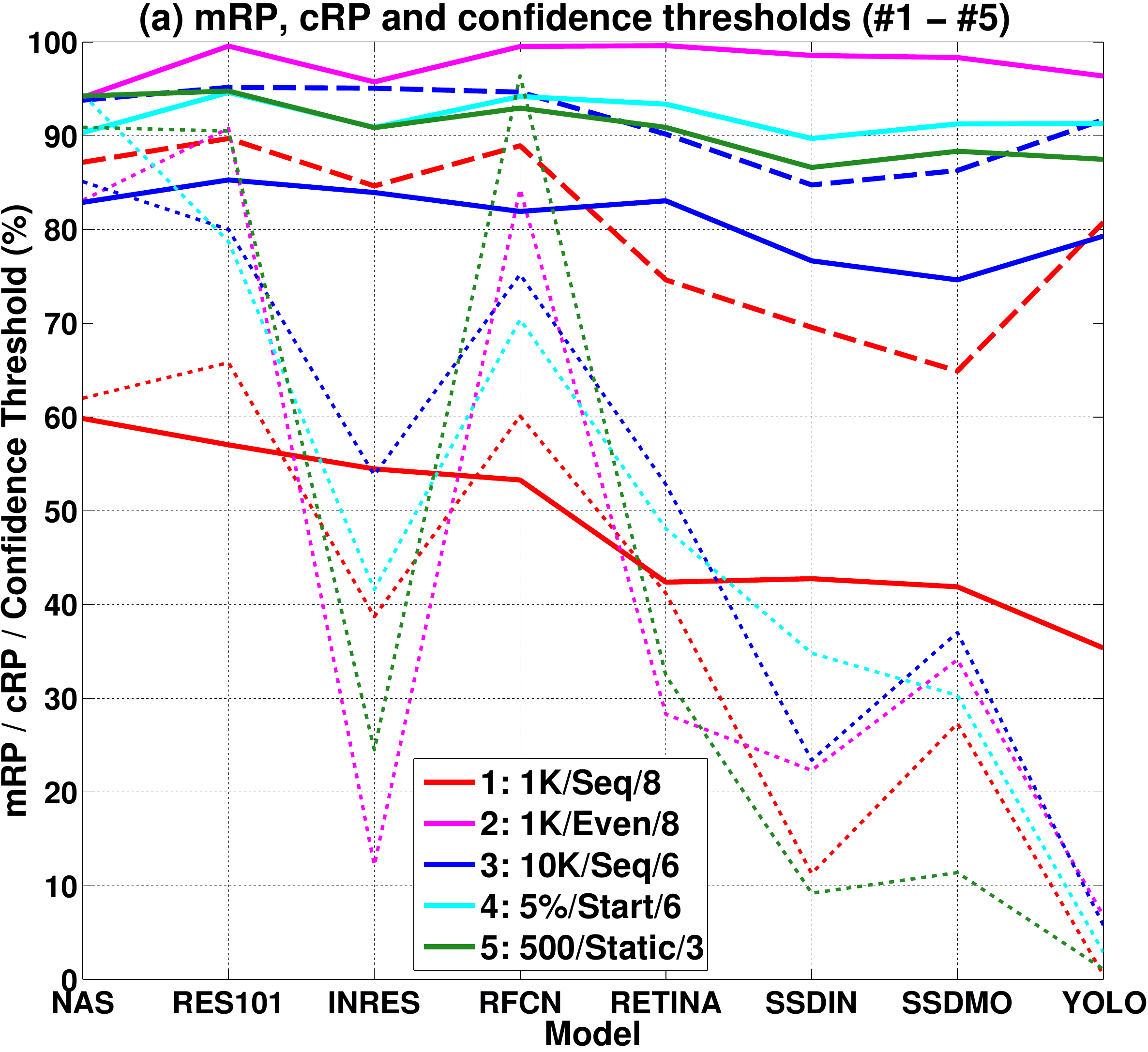}
	\includegraphics[width=0.33\textwidth]{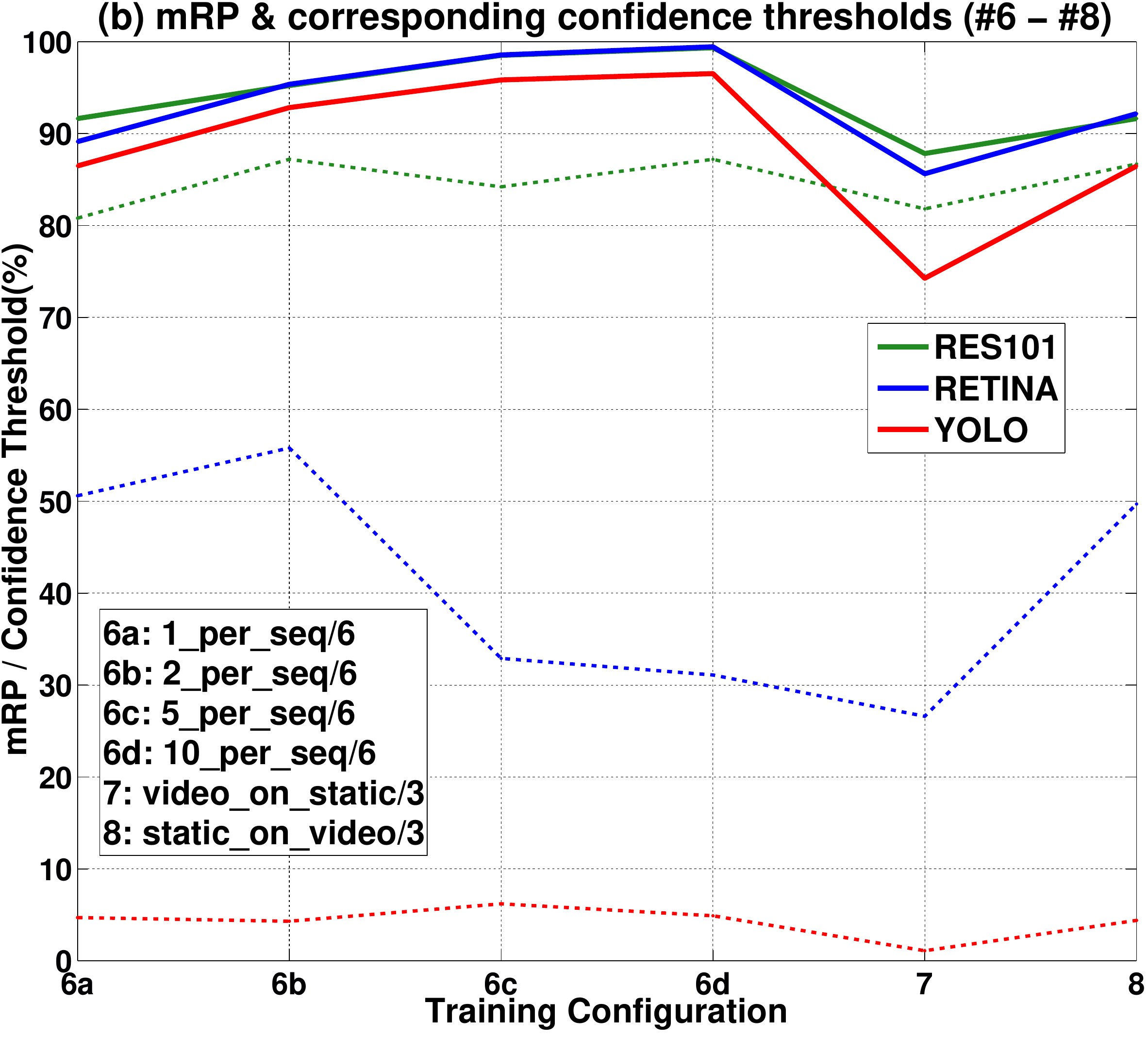}
	\includegraphics[width=0.33\textwidth]{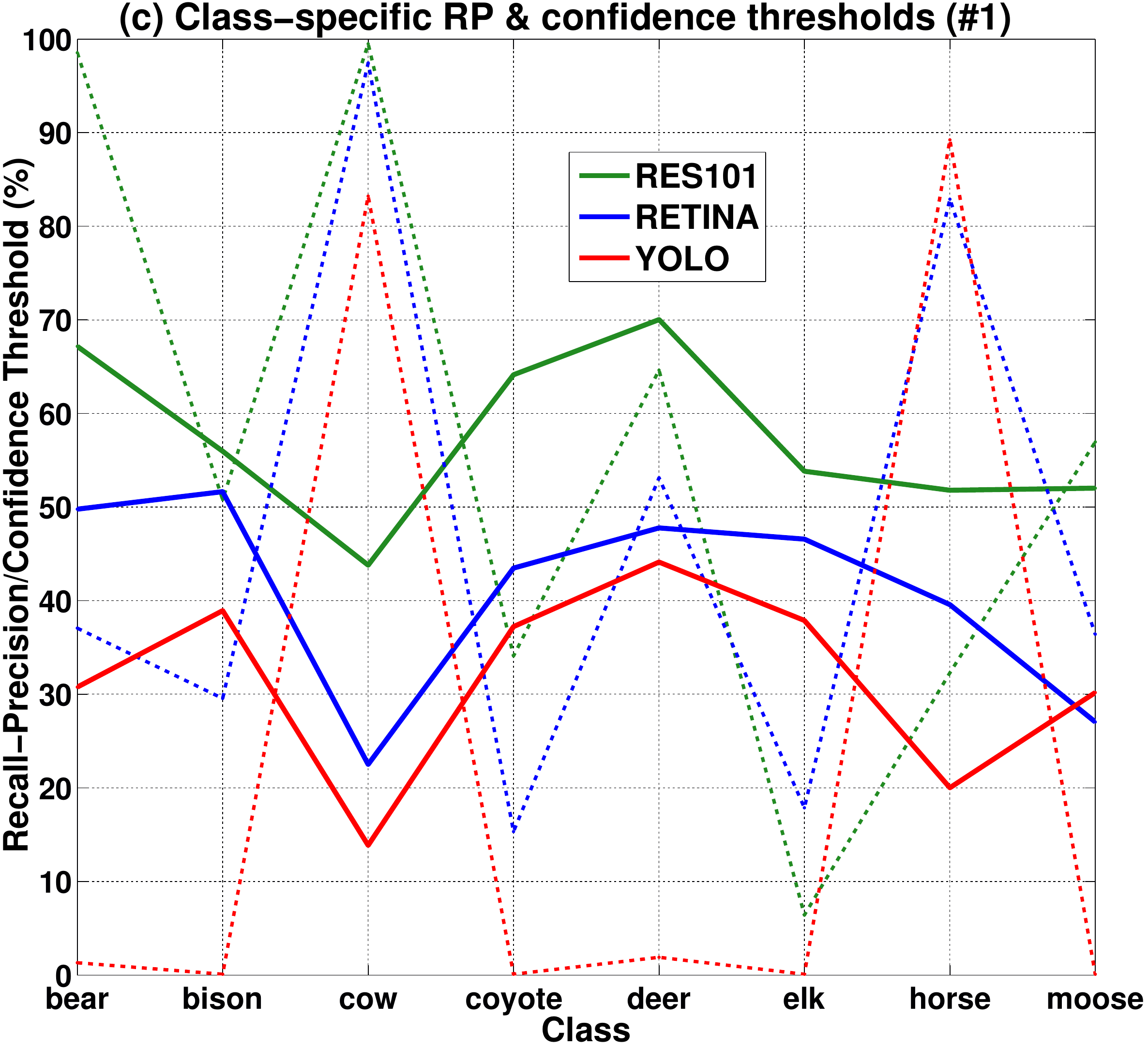}
	\caption{
		Detection mRP (solid), corresponding confidence thresholds (dotted) and cRP (dashed, only \#1, \#3):
		(a) \#1 - \#5 for all 8 models
		(b) \#6 - \#8 for 3 models
		(c) class-specific \#1 results for 3 models.
		Model Acronyms:
		\textbf{NAS}, \textbf{RES101}, \textbf{INRES} - Faster RCNN w/ NAS, ResNet101, Inception-ResNetv2 backbones;
		\textbf{RFCN} - RFCN w/ ResNet101;
		\textbf{RETINA} - RetinaNet w/ ResNet50;
		\textbf{SSDIN}, \textbf{SSDMO} - SSD w/ Inceptionv2, MobileNetv2;
		\textbf{YOLO} - YOLOv3.
		Best viewed under high magnification.
	}
	\label{fig_overview}
\end{figure*}

\subsection{Implementations and Training}
\label{sec_implementations}


Table \ref{tab_implementations} lists all implementations used here.
Training was done by fine tuning models pre-trained on large benchmark datasets – COCO \cite{coco} for Mask RCNN and all detectors; ImageNet \cite{imagenet_cvpr09} for Sharpmask and FCIS; ADE20K \cite{zhou2017_ade20k} for Deeplab, UNet and SegNet.
HED and all trackers were used directly with pretrained weights without any fine tuning.

In order to avoid class bias while training, number of samples from all classes must be similar \cite{Wang16_class_imbalance}.
Number of labeled images, however, varies significantly between animals (Table \ref{tab_annotations}), especially when the source type – video or static
– is taken into account.
Therefore,
experiments were done
with 3, 4 and 6 classes (Table \ref{tab_class_configs}) in addition to all 8 to cover a range of
scenarios while maintaining class balance.

\begin{table}[t]
	\caption{Implementations used for the various methods\\ (TF: Tensorflow, PT: PyTorch)}
	{
	\footnotesize
	\begin{tabular}{|m{4.0cm}|m{3.8cm}|}
		\hline
		\textbf{Methods}                            & \textbf{Implementations}        \\ \hline
		All static detectors except YOLO & TF (Object Detection API) \cite{tf_api_github} \\ \hline
		YOLOv3, YOLOv2, YOLOv1                      & PT \cite{yolov3_github},  TF \cite{darkflow_github}, Darknet \cite{Redmon_darknet}   \\ \hline
		Mask RCNN, Sharpmask, FCIS                    & TF \cite{tf_api_github}, TF \cite{sharpmask_github},  MXNet \cite{fcis_github}                 \\ \hline
		SiamFC,  SiamMask,  DASiamRPN  & TF \cite{SiamFC_github}, PT \cite{SiamMask_github}, PT \cite{DaSiamRPN_github}    \\ \hline
		Deeplab, UNet/SegNet, HED                        & TF \cite{deeplab_repo}, Keras \cite{imgseg_keras}, OpenCV\cite{opencv_library}                \\ \hline
	\end{tabular}
	}
	\label{tab_implementations}
\end{table}

\section{Results}
\label{sec_results}

\subsection{Evaluation Metrics}
\label{sec_metrics}
Object detectors are usually evaluated
using their \textbf{mean average precision (mAP)} \cite{Huang17_tf_api},
defined as the mean,
over all classes,
of the
area under the recall–precision curve
for each class.
Although a good measure of the overall threshold-independent performance, mAP may not accurately represent deployment accuracy where a single threshold must be chosen.
Since mAP considers the variation of recall and precision with threshold \textit{separately for each class}, and this can differ greatly between classes (Fig. \ref{fig_overview}c), it is more indicative of accuracy when a different threshold can be chosen for each class to optimize the recall-precision characteristics for that class.
It is also difficult to interpret mAP to gauge the practical usability of a detector in terms of how likely it is to miss objects or give false detections.
This paper therefore proposes another metric obtained by first
averaging
recall and precision for each threshold over all classes and then taking the \textbf{recall–precision (RP)} value at the threshold where the two are equal.
This metric is named \textbf{mean Recall-Precision (mRP)} and provides a more interpretable measure of performance when using a single threshold for all classes.

Further, this work deals mainly with
human-in-the-loop type
security applications where detections
alert humans to take suitable measures after verification.
In such cases, simply detecting an object can be far more crucial than classifying it correctly.
For example, when used as an early warning system for bus drivers, misclassification would have little impact on the driver's response as long as the animal is detected early enough.
A less stringent evaluation criterion named \textbf{class-agnostic Recall-Precision (cRP)} is thus also used that treats all animals 
as
belonging to
the same class
so that misclassifications are not penalized.

\subsection{Real Data} 
\label{sec_res_real}

\subsubsection{How well do detectors generalize ?}
\label{sec_res_generalize}

Fig. \ref{fig_overview} summarizes the results for several training and testing configurations (Table \ref{tab_train_configs}) used to study the generalization ability of detectors in a range of scenarios.
These are henceforth referred to by their \textbf{numeric IDs} (first column of Table \ref{tab_train_configs}) and detectors by \textbf{acronyms} (Fig. \ref{fig_overview}) for brevity.

Fig. \ref{fig_overview}a gives results for all detectors in \#1 - \#5.
The large difference between \#1 and \#2 clearly demonstrates the inability of detectors to generalize to unseen backgrounds.
Both have 1K video images/class but the latter has these sampled from all sequences to allow training over nearly all backgrounds in the test set while the former
does not get any frames from the tested sequences.
This is sufficient for the detectors to achieve near perfect mRPs in \#2 while giving
far poorer performance with
only 35-60\% mRP in \#2.
A similar trend is seen, though to a lesser extent, in \#3 and \#4.
The former, with 10K images/class from complete sequences, is significantly outperformed by the latter with only 5\% images from the start of each sequence (or $\sim$1.2K images/class).
The smaller difference here is attributable to the much greater frame count in \#3 and the fact that \#4 uses consecutive frames from each sequence which contain a smaller range of backgrounds than the evenly sampled frames in \#2.
Performance in \#5
is comparable to \#4, even though \#5 involves testing over a far greater proportion of unseen backgrounds,
probably because most static images depict animals in their natural habitats 
(Sec. \ref{sec_data_collection})
which, exhibiting limited variability, allow the detectors to generalize relatively well.

Fig. \ref{fig_overview}a also shows cRP, though only for \#1 and \#3 since remaining configurations all had cRP > 90\% whose inclusion would have cluttered the plots so these 
have been
deferred to the supplementary. 
As expected, cRPs are significantly higher than mRPs for all models, though the gain is most notable for YOLO, particularly in \#1 where it more than doubles
its performance,
outperforming both the SSDs as well as RETINA.
This suggests, and qualitative examination has confirmed, that the form of overfitting YOLO is susceptible to involves associating backgrounds to specific animals whose training images had similar backgrounds.
For example, if a particular scene is present in bear training images but not in those of deer, a test image of a similar scene, but containing deer, would have the animal detected as bear.
The other models are susceptible to this too but to a smaller degree and more often miss the animal altogether.

\begin{table}[t]
	\centering
	\caption{Class configurations for training
		(\textbf{c}: no. of classes)
	}
	{
		\footnotesize
		\begin{tabular}{|m{0.15cm}|m{2.9cm}|m{3.4cm}|}
			\hline
			\textbf{c} & \textbf{Animals}          & \textbf{Comments}                                           \\ \hline
			6                   & all except cow, horse      & these have only $\sim$5K images \\ \hline
			4                   & bear, deer, moose, coyote & synthetic images                                     \\ \hline
			3                   & bear, deer, coyote        & real static images                                          \\ \hline
		\end{tabular}
	}
	\label{tab_class_configs}
\end{table}

\begin{table}[t]
	\centering
	\caption{
		Training configurations for both real and synthetic data
		(\textbf{c}, \textbf{img}, \textbf{seq}
		-
		number of classes, images, sequences).
	}
	{
		\footnotesize
		\begin{tabular}{|m{0.3cm}|m{0.1cm}|m{3.8cm}|m{1.1cm}|m{0.9cm}|}
			\hline
			\multicolumn{1}{|c|}{\multirow{2}{*}{\textbf{\#}}} & \multicolumn{1}{c|}{\multirow{2}{*}{\textbf{c}}} & \multicolumn{1}{c|}{\multirow{2}{*}{\textbf{Details}}}           & \multicolumn{1}{c|}{\textbf{Train}} & \multicolumn{1}{c|}{\textbf{Test}}  \\ \cline{4-5} 
			\multicolumn{1}{|c|}{}                             & \multicolumn{1}{c|}{}                              & \multicolumn{1}{c|}{}                                            & \multicolumn{1}{c|}{\textbf{img (seq)}}    & \multicolumn{1}{c|}{\textbf{img (seq)}} \\ \hline
			\textbf{1}                                        & 8                                                  & 1K video images/class sampled from complete sequences                   & 8001 (33)                                  & 150117 (539)                            \\ \hline
			\textbf{2}                                         & 8                                                  & 1K video images/class sampled evenly across all sequences           & 8156                                       & 149962                                 \\ \hline
			\textbf{3}                                         & 6                                                  & 10K video images/class sampled from complete sequences                & 60003 (218)                                & 88023 (317)                             \\ \hline
			\textbf{4}                                       & 6                                                  & 5\% images from the start of each video sequence & 7169                                       & 140857                                 \\ \hline
			\textbf{5}                                         & 3                                                  & 500 static images/class                                                & 1500                                       & 2900                                    \\ \hline
			\textbf{6a}-\textbf{6d}                                       &     6                                             & 1, 2, 5, 10 images sampled evenly from each of 67 sequences & 402, 804, 2010,4020                                       & 103235                                 \\ \hline
			\textbf{7}                                         & 3                                                  & 20K video images/class tested on static images                                               &        60000                                & 4400                                    \\ \hline
			\textbf{8a, 8b}                                         & 3                                                  & 1K static images/class tested on video, synthetic images                                            &        3000                                & 73034, 598                                   \\ \hline
			\textbf{9}                                         & 4                                                  & 20K video images/class tested on synthetic images                                               &        80008                                & 780                                    \\ \hline
			
			\textbf{10a, 10b}                                         & 3, 4                                                  & 3, 4 class models trained on 28\% of synthetic images, tested on rest                                               &        234, 312                                & 598, 780                                    \\ \hline
		\end{tabular}
	}
	\label{tab_train_configs}
\end{table}

\subsubsection{How much are video annotations worth ?}
\label{sec_res_vid_worth}
Fig. \ref{fig_overview}b shows results for \#6 - \#8; only 3 detectors are included to reduce clutter since the others showed similar performance patterns.
\#6 involved training with 1, 2, 5 and 10 frames/sequence, with the sequence count limited to 67 by the class with the fewest sequences (deer) to maintain class balance.
All 4 models were tested on the same 67 sequences using frames not included in any of their training sets.
It can be seen that even 1 frame/sequence is enough for all detectors to give $~90\%$ mRP, which improves only marginally with 2 and 5 frames, plateauing thereafter.
Further, though RETINA does catch up with RES101 using $ \geq 2 $ frames, YOLO is unable to even narrow the gap,
which might indicate that domain specialization cannot entirely overcome architectural limitations.
\#7 and \#8 show the relative utility of video and static frames by training on one and testing on the other.
As expected, static models demonstrate far superior generalizability by outperforming the video models by 4-12\% mRP even though the latter are trained and tested on $ 20\times $ more and $ 16\times $ fewer frames respectively.
Performance gap between \#7 and \#8 is also larger for worse performing models, especially YOLO that has twice the gap of RETINA, which reaffirms its poor generalizability.
Finally, the fact that \#8 has lower mRP than \#6a even though the former has nearly $ 15\times $ more images
with varied backgrounds
shows the importance of domain specialization. 
\subsubsection{How do the detector accuracies compare ?}
\label{sec_res_compare}
RES101 turns out to be the best overall, though NAS, RFCN and INRES remain comparable in all configurations.
NAS even has a slight edge in \#1, showing its better generalizability under the most difficult scenarios.
Conversely, the shortcomings of 1-stage detectors compared to their 2-stage counterparts are also most apparent in \#1.
This is particularly notable for RETINA that is comparable to RES101 and significantly better than the other 1-stage detectors in all remaining configs.
YOLO likewise performs much poorer relative to the two SSDs
while being similar and even better in other configs. 
This might indicate that 1-stage detectors in general, and YOLO in particular, are more prone to overfitting with limited training data.
From a practical standpoint, though, YOLO redeems itself well by its relatively high cRPs, outperforming RETINA
in both \#1 and \#3.
\subsubsection{How important is the confidence threshold ?}
\label{sec_res_thresh}
Fig. \ref{fig_overview} shows confidence thresholds corresponding to mRP or class-specific RP using dotted lines.
Fig. \ref{fig_overview}c shows that
the threshold corresponding to the class-specific RP varies widely between classes - much more than the RP itself.
As mentioned in Sec. \ref{sec_metrics}, this motivates the use of mRP instead of mAP as a 
practical
evaluation criterion.
Further, Fig. \ref{fig_overview}a,b show that the optimal mRP threshold itself varies greatly between the detectors too.
Therefore, choosing a single threshold for all of them
might not provide a true picture of their relative performance in practice.
It is also evident, especially in Fig. \ref{fig_overview}b, that a weak correlation exists between the relative performance and threshold, with better performing detectors usually also having higher thresholds.
Notable exceptions to this are INRES  and SSDIN, both having smaller thresholds than their respective mRP levels.
Since both use different variants of Inception, this might be due to an architectural peculiarity thereof.
Also notable are the very low thresholds of YOLO - often  $ <5\% $ and sometimes even $ <1\% $.

\subsubsection{How resource intensive are the detectors ?}
\label{sec_res_speed}

\begin{table}[t]
	\centering
	\caption{
		Speed, GPU memory consumption and maximum batch size for each detector.
		Refer Fig. \ref{fig_overview} for model names.
		(Setup: Titan Xp 12GB, Threadripper 1900X, 32GB RAM)
	}
{
\footnotesize
	\begin{tabular}{|m{0.9cm}|m{0.8cm}|m{0.7cm}|m{0.8cm}|m{0.7cm}|m{0.8cm}|m{0.7cm}|}
		\hline
		\multirow{2}{*}{\textbf{Model}} & \multicolumn{2}{l|}{\textbf{Batch Size 1}}  & \multicolumn{2}{l|}{\textbf{Batch Size 4}}  & \multicolumn{2}{l|}{\textbf{Max Batch Size}} \\ \cline{2-7} 
		& \textbf{memory (MB)} & \textbf{speed (FPS)} & \textbf{memory (MB)} & \textbf{speed (FPS)} & \textbf{batch size}    & \textbf{speed (FPS)}    \\ \hline
		\textbf{NAS}                    & 9687                 & 1.36                 & -                    & -                    & 3                      & 1.39                    \\ \hline
		\textbf{INRES}               & 7889                 & 3.95                 & 8145                 & 4.68                 & 8                      & 4.49                    \\ \hline
		\textbf{RES101}                 & 5077                 & 19.61                & 5589                 & 25.35                & 36                     & 27.12                   \\ \hline
		\textbf{RFCN}                   & 5041                 & 19.8                 & 5553                 & 32.12                & 76                     & 26.94                   \\ \hline
		\textbf{RETINA}                 & 4785                 & 31.5                 & 5553                 & 43.51                & 120                    & 53.28                   \\ \hline
		\textbf{YOLO}                   & 1487                 & 71.41                & 2039                 & 104.25               & 48                     & 119.64                  \\ \hline
		\textbf{SSDIN}               & 3631                 & 68.35                & 3631                 & 155.63               & 160                    & 181.66                  \\ \hline
		\textbf{SSDMO}                & 1999                 & 78.67                & 2031                 & 167                  & 480                    & 246.56                  \\ \hline
	\end{tabular}
}
	\label{tab_speeds}
\end{table}

\begin{figure*}[!htbp]
	\includegraphics[width=0.33\textwidth]{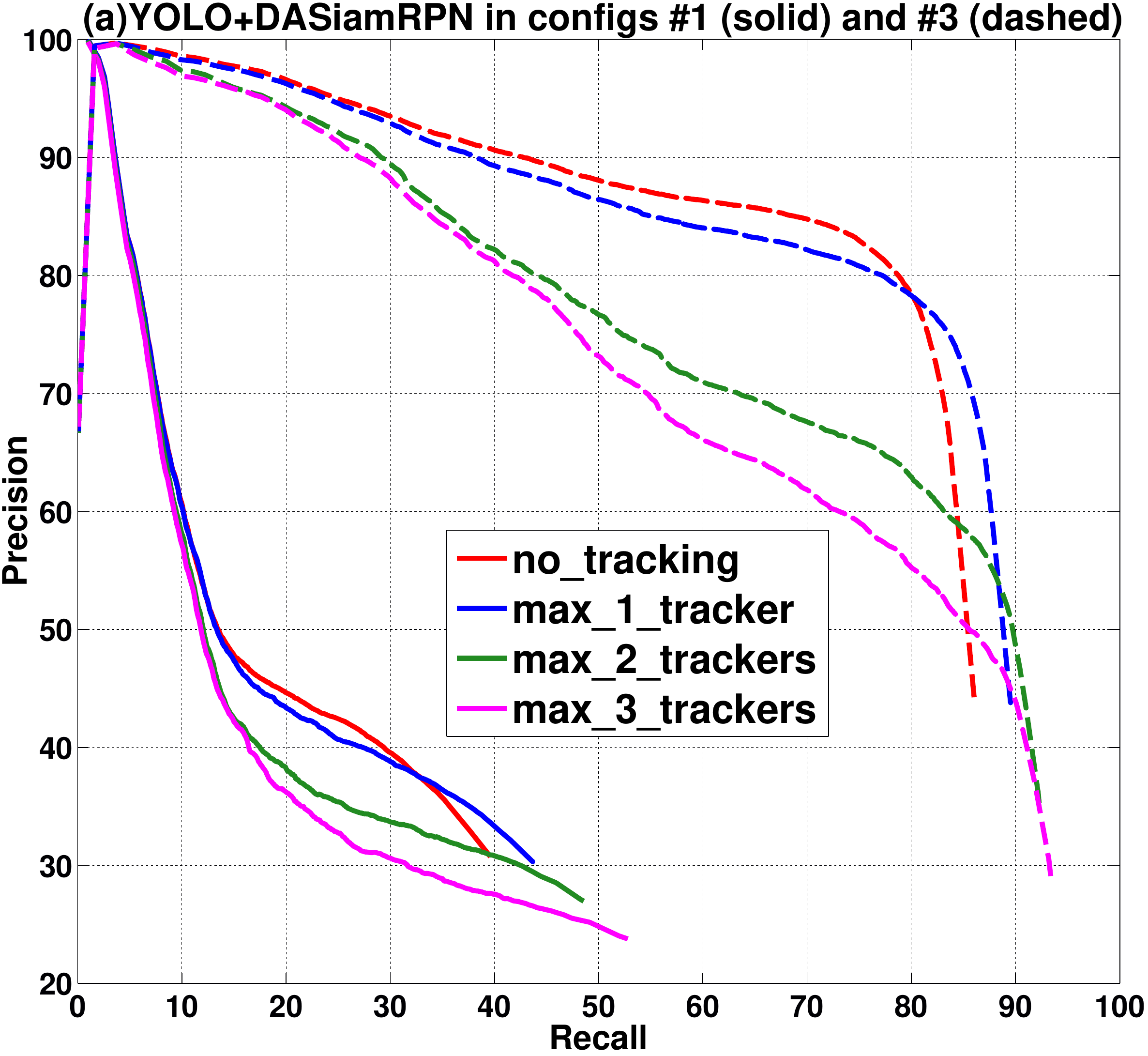}
	\includegraphics[width=0.33\textwidth]{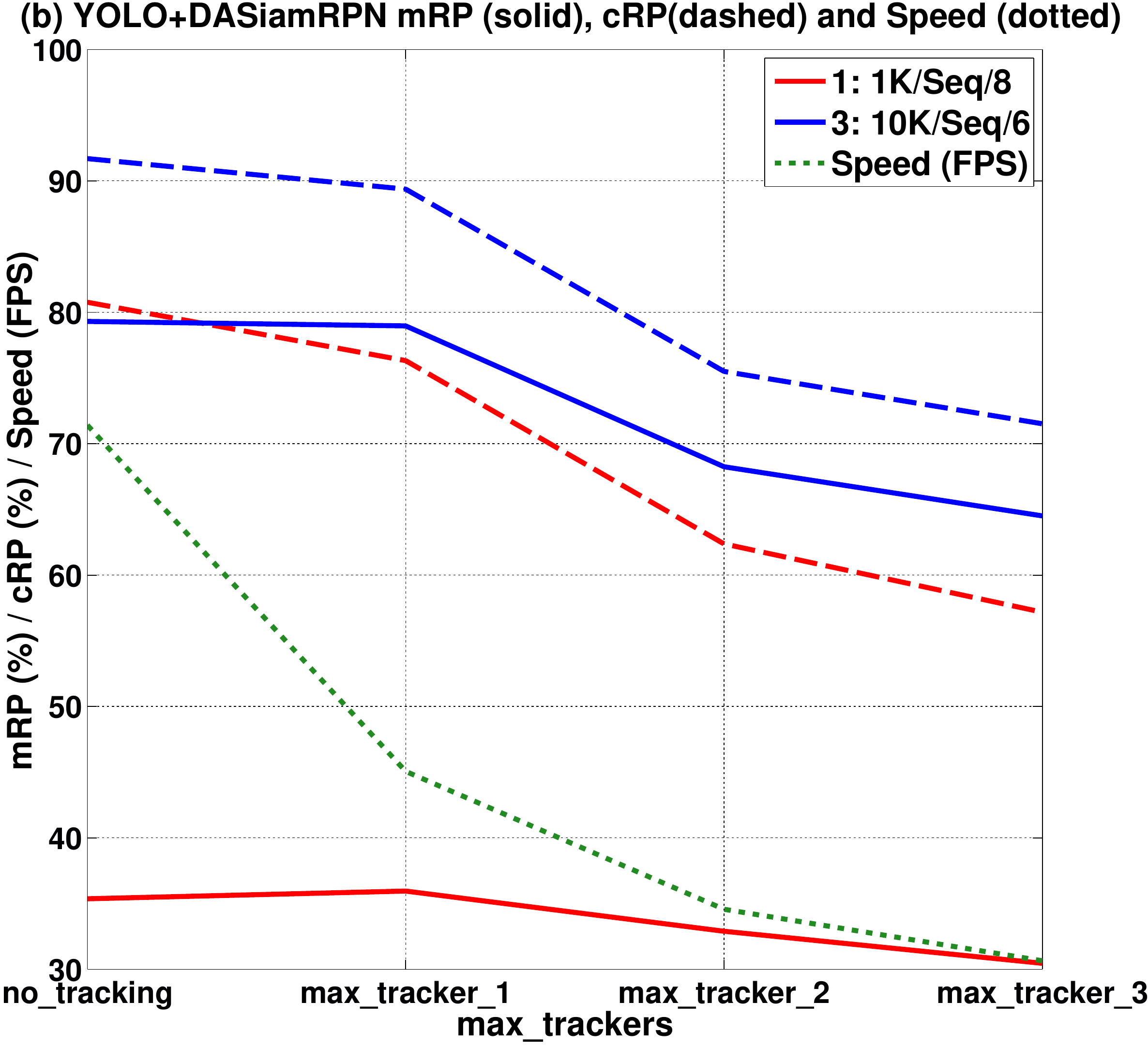}
	\includegraphics[width=0.33\textwidth]{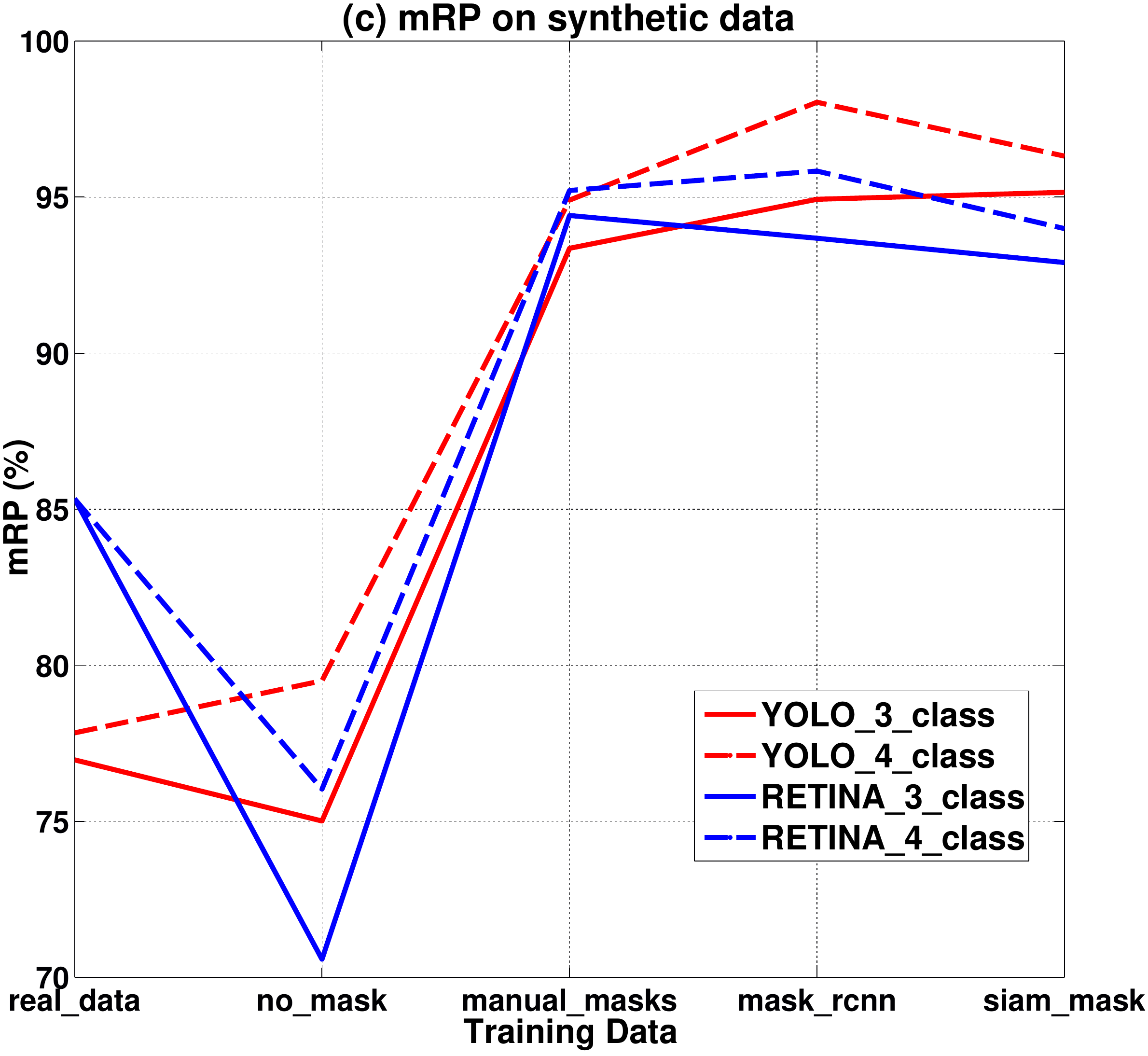}
	\caption{
		Results for (a - b) DASiamRPN + YOLO and (c) RETINA and YOLO tested on synthetic data
	}
	\label{fig_details}
\end{figure*}


Since both deployment scenarios of ATV and highway buses involve mobile systems with limited power availability, it is important for the detector to be
as lightweight as possible.
Table \ref{tab_speeds} shows the speed in frames per second (FPS) along with GPU memory consumption for batch sizes 1 and 4, where the latter is chosen to represent the 4 cameras needed for a simultaneous 360° field-of-view.
The maximum batch size that can be run on a 12GB Titan Xp GPU is also shown for scalability comparison.
SSDMO turns out to be the fastest, though YOLO is comparable at batch size 1 and also has significantly smaller memory footprint.
However, YOLO does not scale as well in either speed or memory and ends up with only a tenth of the maximum batch size of SSDMO and less than half the corresponding speed.
NAS and INRES are the slowest and most memory intensive by far and unsuitable for realtime applications.
RFCN and RES101 are similar
with unit batch size, probably due to their identical backbone, though RFCN scales better, allowing more than twice the maximum batch size and 28\% higher speed with batch size 4.
Finally, RETINA provides the best compromise between performance and speed - RES101-like mRP in most configs and fast enough to process 4 camera streams simultaneously at 10 FPS each and thus
capture an animal visible for
a fraction of
a second.

\vspace{-0.35cm}

\subsubsection{Can tracking reduce false negatives ?}
\label{sec_res_tracking}

As mentioned in Sec. \ref{sec_detection}, tracking was used in an attempt to reduce false negatives by utilizing temporal information in videos.
DASiamRPN \cite{Zhu18_dasiamrpn} was used
as the tracker
as being one of the fastest available Siamese type trackers.
YOLO was used as the detector since its PyTorch implementation
was easier to integrate with that of DASiamRPN,
its speed with batch size 1 (necessary to use tracking) is among the fastest and its poor performance in \#1 provides ample scope for improvement.
The detailed algorithm is included in the supplementary, though its high level idea is simple - associate detections with existing trackers, create new trackers for unassociated detections and remove trackers that remain unassociated for too long or those with the lowest confidence when tracker count exceeds a threshold.
Fig. \ref{fig_details}a shows the mean Recall vs. Precision plots while
Fig. \ref{fig_details}b
gives mRP / cRP and speeds.
Tracking mostly helps only when the detector finds an animal in at least one frame in a sequence and misses it in several subsequent ones.
It turns out that this seldom happens
in practice so that the resultant increase in recall is very slight and is offset by a significant decrease in precision through tracking false positives.
The latter can be mitigated by removing unassociated trackers frequently but this leads to a large drop in recall and is therefore not used in these results.
There is thus no net gain in mRP/cRP using tracking, rather significant drops with >1 trackers.
When combined with the reduction in FPS, it does not seem like an effective way to reduce false negatives.
\vspace{-0.4cm}

\subsubsection{Can multi-model pooling reduce false negatives ?}
\label{sec_polling}
Another way to reduce missing detections is to run multiple detectors simultaneously and pool their detections.
A large variety of methods were explored to pool
YOLO, SSDIN and SSDMO
but none
managed to increase
recall enough to offset the fall in precision and the net mRPs were even worse than those from tracking.
Descriptions of these methods
and
corresponding results
are thus in
the supplementary.

\vspace{-0.1cm}

\subsection{Synthetic data}
\label{sec_res_synthetic}
A training set was constructed from synthetic data by selecting images corresponding to 3 animal poses per background, with a different combination of poses selected randomly for each background, while all remaining images were used for testing.
Table \ref{tab_train_configs} denotes the corresponding configs as \#10a and \#10b for 3 and 4 class models respectively. 
Corresponding real data configurations are \#8b  and \#9  with 1K static and 20K video images/class respectively.
Seperate models were trained for each of the 4 methods of extracting animals from source images (Sec. \ref{sec_synthetic}) – Gaussian blending, manual masks, Mask RCNN and SiamMask.
All were tested on images generated by manual masks.

As shown in Fig. \ref{fig_details}c, models trained on synthetic data significantly outperform those trained on real data as long as masks are used.
This is remarkable considering that only 78 frames/class were used for the former compared to 1K or 20K for the latter.
This reiterates the results in Sec. \ref{sec_res_vid_worth} where \#6a with 67 images outperformed \#8a with the same 1K images as \#8b.
However, unlike there, YOLO does manage to match RETINA here, which suggests that high enough degree of specialization can indeed overcome its architectural shortcomings.
More importantly, there is no perceptible difference in mRP between models corresponding to the three segmentation methods.
This shows that even the fully unsupervised and visibly coarse masks from SiamMask have comparable detector training information to precise manual masks.
At the same time, mask quality does indeed matter since the no mask / Gaussian blending models perform even worse than real data.

\vspace{-0.28cm}

\section{Conclusions}
\label{sec_conclusions}
This paper presented a large scale study of animal detection with deep learning where 8 state of the art detectors were compared in a wide range of configurations.
A particular focus of the study was to evaluate their generalization ability when
training and test scenarios
do not match.
It was shown that none of the detectors can generalize well enough to provide usable models for deployment, with missed detections on previously unseen backgrounds being the main issue.
Attempts to increase recall using tracking and multi-model pooling proved ineffective.
Synthetic data generation using segmentation masks to extract animals from images of natural habitats and inserting them in target scenes was shown to be an effective solution.
An almost fully automated way to achieve this was demonstrated by the competitiveness of coarse unsupervised masks with precise manual ones in terms of the performance of detectors trained on the corresponding synthetic images.
RETINA and YOLO were shown to be competitive with larger models while being sufficiently lightweight for multi-camera mobile deployment.

{\small
\bibliographystyle{ieee}
\bibliography{references}

\begin{thebibliography}{100}\itemsep=-1pt

\bibitem{coco}
{COCO Dataset}.
\newblock online: \url{http://cocodataset.org/}, September 2018.

\bibitem{fcis_github}
{Fully Convolutional Instance-aware Semantic Segmentation}.
\newblock online: \url{https://github.com/msracver/FCIS}, September 2019.

\bibitem{Labelbox}
Labelbox.
\newblock online: https://labelbox.com/, 2019.

\bibitem{lilabc19_ecology_dataset}
Labeled information library of alexandria: Biology and conservation.
\newblock online: http://lila.science/datasets1, August 2019.

\bibitem{nature_footage}
Nature footage.
\newblock \url{https://www.naturefootage.com}, 2019.

\bibitem{playment}
Playment.
\newblock online: https://playment.io, 2019.

\bibitem{sharpmask_github}
{TensorFlow implementation of DeepMask and SharpMask}.
\newblock online: \url{https://github.com/aby2s/sharpmask}, September 2019.

\bibitem{lstm_ssd19}
{Tensorflow Mobile Video Object Detection}.
\newblock Github, 2019.
\newblock
  \url{https://github.com/tensorflow/models/research/lstm_object_detection}.

\bibitem{tf_api_github}
{Tensorflow Object Detection API}.
\newblock online:
  \url{https://github.com/tensorflow/models/tree/master/research/object_detection},
  Sept 2019.

\bibitem{animal_detection_github}
{Deep Learning for Animal Detection in Man-made Environments}.
\newblock online: \url{https://github.com/abhineet123/animal_detection},
  January 2020.

\bibitem{BadrinarayananK17_segnet}
V.~Badrinarayanan, A.~Kendall, and R.~Cipolla.
\newblock {SegNet: A Deep Convolutional Encoder-Decoder Architecture for Image
  Segmentation}.
\newblock {\em {IEEE} Trans. Pattern Anal. Mach. Intell.}, 39(12):2481--2495,
  2017.

\bibitem{beery2019iwildcam}
S.~Beery, D.~Morris, and P.~Perona.
\newblock {The iWildCam 2019 Challenge Dataset}.
\newblock {\em arXiv preprint arXiv:1907.07617}, 2019.

\bibitem{Beery2018_caltech_trap_dataset}
S.~Beery, G.~Van~Horn, and P.~Perona.
\newblock {Recognition in Terra Incognita}.
\newblock In V.~Ferrari, M.~Hebert, C.~Sminchisescu, and Y.~Weiss, editors,
  {\em Computer Vision -- ECCV 2018}, pages 472--489, Cham, 2018. Springer
  International Publishing.

\bibitem{Sara18_trap}
S.~Beery, G.~Van~Horn, and P.~Perona.
\newblock Recognition in terra incognita.
\newblock In V.~Ferrari, M.~Hebert, C.~Sminchisescu, and Y.~Weiss, editors,
  {\em Computer Vision -- ECCV 2018}, pages 472--489, Cham, 2018. Springer
  International Publishing.

\bibitem{Benenson2019LargescaleIO}
R.~Benenson, S.~Popov, and V.~Ferrari.
\newblock {Large-scale interactive object segmentation with human annotators}.
\newblock {\em CVPR}, abs/1903.10830, 2019.

\bibitem{SiamFC_github}
L.~Bertinetto and J.~Valmadre.
\newblock {SiamFC - TensorFlow}.
\newblock online: \url{https://github.com/torrvision/siamfc-tf}, September
  2019.

\bibitem{Luca16_siamfc}
L.~Bertinetto, J.~Valmadre, J.~F. Henriques, A.~Vedaldi, and P.~H.~S. Torr.
\newblock {Fully-Convolutional Siamese Networks for Object Tracking}.
\newblock In G.~Hua and H.~J{\'e}gou, editors, {\em Computer Vision -- ECCV
  2016 Workshops}, pages 850--865, Cham, 2016. Springer International
  Publishing.

\bibitem{opencv_library}
G.~Bradski.
\newblock {The OpenCV Library}.
\newblock {\em Dr. Dobb's Journal of Software Tools}, 2000.

\bibitem{Breheret17_annotation}
A.~Br{\'e}h{\'e}ret.
\newblock {Pixel Annotation Tool}.
\newblock \url{https://github.com/abreheret/PixelAnnotationTool}, 2017.

\bibitem{Chen14_trap}
G.~{Chen}, T.~X. {Han}, Z.~{He}, R.~{Kays}, and T.~{Forrester}.
\newblock Deep convolutional neural network based species recognition for wild
  animal monitoring.
\newblock In {\em 2014 IEEE International Conference on Image Processing
  (ICIP)}, pages 858--862, Oct 2014.

\bibitem{ChenPKMY18_deeplab}
L.~Chen, G.~Papandreou, I.~Kokkinos, K.~Murphy, and A.~L. Yuille.
\newblock Deeplab: Semantic image segmentation with deep convolutional nets,
  atrous convolution, and fully connected crfs.
\newblock {\em {IEEE} Trans. Pattern Anal. Mach. Intell.}, 40(4):834--848,
  2018.

\bibitem{deeplab_repo}
L.-C. Chen, Y.~Zhu, and G.~Papandreou.
\newblock {DeepLab: Deep Labelling for Semantic Image Segmentation}.
\newblock Github, 2017.
\newblock
  \url{hhttps://github.com/tensorflow/models/tree/master/research/deeplab}.

\bibitem{deeplabv3plus2018}
L.-C. Chen, Y.~Zhu, G.~Papandreou, F.~Schroff, and H.~Adam.
\newblock Encoder-decoder with atrous separable convolution for semantic image
  segmentation.
\newblock {\em arXiv:1802.02611}, 2018.

\bibitem{Chollet17_xception}
F.~Chollet.
\newblock Xception: Deep learning with depthwise separable convolutions.
\newblock In {\em {CVPR}}, pages 1800--1807. {IEEE} Computer Society, 2017.

\bibitem{opt_transport13}
M.~Cuturi.
\newblock Sinkhorn distances: Lightspeed computation of optimal transport.
\newblock In C.~J.~C. Burges, L.~Bottou, M.~Welling, Z.~Ghahramani, and K.~Q.
  Weinberger, editors, {\em NIPS}, pages 2292--2300. Curran Associates, Inc.,
  2013.

\bibitem{Dai16_rfcn}
J.~Dai, Y.~Li, K.~He, and J.~Sun.
\newblock R-fcn: Object detection via region-based fully convolutional
  networks.
\newblock In D.~D. Lee, M.~Sugiyama, U.~V. Luxburg, I.~Guyon, and R.~Garnett,
  editors, {\em Advances in Neural Information Processing Systems 29}, pages
  379--387. Curran Associates, Inc., 2016.

\bibitem{imagenet_cvpr09}
J.~Deng, W.~Dong, R.~Socher, L.-J. Li, K.~Li, and L.~Fei-Fei.
\newblock {ImageNet: A Large-Scale Hierarchical Image Database}.
\newblock In {\em CVPR}, 2009.

\bibitem{dutta2016via}
A.~Dutta, A.~Gupta, and A.~Zissermann.
\newblock {VGG} image annotator ({VIA}).
\newblock http://www.robots.ox.ac.uk/~vgg/software/via/, 2016.

\bibitem{dutta2019vgg}
A.~Dutta and A.~Zisserman.
\newblock The {VIA} annotation software for images, audio and video.
\newblock In {\em Proceedings of the 27th ACM International Conference on
  Multimedia}, MM '19, New York, NY, USA, 2019. ACM.

\bibitem{Everingham10_voc}
M.~Everingham, L.~Van~Gool, C.~K.~I. Williams, J.~Winn, and A.~Zisserman.
\newblock The pascal visual object classes (voc) challenge.
\newblock {\em International Journal of Computer Vision}, 88(2):303--338, June
  2010.

\bibitem{Feichtenhofer17_dt}
C.~Feichtenhofer, A.~Pinz, and A.~Zisserman.
\newblock Detect to track and track to detect.
\newblock In {\em ICCV}, 2017.

\bibitem{Forslund14_vid_night}
D.~{Forslund} and J.~{Bjarkefur}.
\newblock Night vision animal detection.
\newblock In {\em 2014 IEEE Intelligent Vehicles Symposium Proceedings}, pages
  737--742, June 2014.

\bibitem{Geiger2013IJRR_kitti}
A.~Geiger, P.~Lenz, C.~Stiller, and R.~Urtasun.
\newblock Vision meets robotics: The kitti dataset.
\newblock {\em International Journal of Robotics Research (IJRR)}, 2013.

\bibitem{Girshick15_fast_rcnn}
R.~{Girshick}.
\newblock Fast r-cnn.
\newblock In {\em ICCV}, pages 1440--1448, Dec 2015.

\bibitem{Girshick14_rcnn}
R.~B. Girshick, J.~Donahue, T.~Darrell, and J.~Malik.
\newblock Rich feature hierarchies for accurate object detection and semantic
  segmentation.
\newblock {\em CVPR}, pages 580--587, 2014.

\bibitem{Girshick2016_rcnn}
R.~B. Girshick, J.~Donahue, T.~Darrell, and J.~Malik.
\newblock Region-based convolutional networks for accurate object detection and
  segmentation.
\newblock {\em IEEE Transactions on Pattern Analysis and Machine Intelligence},
  38:142--158, 2016.

\bibitem{Grace2017_avc}
M.~K. Grace, D.~J. Smith, and R.~F. Noss.
\newblock Reducing the threat of wildlife-vehicle collisions during peak
  tourism periods using a roadside animal detection system.
\newblock {\em Accident Analysis \& Prevention}, 109:55--61, 2017.

\bibitem{gray19_deep_ecology}
P.~Gray.
\newblock Awesome deep ecology.
\newblock online: {\url{https://github.com/patrickcgray/awesome-deep-ecology}},
  Month = {August}, Year = {2019}, Owner = {Tommy}, Timestamp = {2018.02.02}.

\bibitem{imgseg_keras}
D.~Gupta.
\newblock {Image Segmentation Keras : Implementation of Segnet, FCN, UNet and
  other models in Keras}.
\newblock Github, 2017.
\newblock \url{https://github.com/divamgupta/image-segmentation-keras}.

\bibitem{Han16_seqnms}
W.~Han, P.~Khorrami, T.~L. Paine, P.~Ramachandran, M.~Babaeizadeh, H.~Shi,
  J.~Li, S.~Yan, and T.~S. Huang.
\newblock Seq-nms for video object detection.
\newblock {\em CoRR}, abs/1602.08465, 2016.

\bibitem{He17_mask_rcnn}
K.~He, G.~Gkioxari, P.~Doll{\'{a}}r, and R.~B. Girshick.
\newblock Mask {R-CNN}.
\newblock In {\em ICCV}, pages 2980--2988, 2017.

\bibitem{He16_resnet}
K.~{He}, X.~{Zhang}, S.~{Ren}, and J.~{Sun}.
\newblock Deep residual learning for image recognition.
\newblock In {\em CVPR}, pages 770--778, June 2016.

\bibitem{He16_resnet2}
K.~He, X.~Zhang, S.~Ren, and J.~Sun.
\newblock Identity mappings in deep residual networks.
\newblock {\em CoRR}, abs/1603.05027, 2016.

\bibitem{Horn19_nabirds_dataset}
G.~V. Horn.
\newblock Nabirds dataset.
\newblock online:\url{http://dl.allaboutbirds.org/nabirds1}, August 2019.

\bibitem{Horn18_inaturalist}
G.~V. {Horn}, O.~M. {Aodha}, Y.~{Song}, Y.~{Cui}, C.~{Sun}, A.~{Shepard},
  H.~{Adam}, P.~{Perona}, and S.~{Belongie}.
\newblock The inaturalist species classification and detection dataset.
\newblock In {\em CVPR}, pages 8769--8778, June 2018.

\bibitem{Howard19_mobilenetv3}
A.~Howard, M.~Sandler, G.~Chu, L.~Chen, B.~Chen, M.~Tan, W.~Wang, Y.~Zhu,
  R.~Pang, V.~Vasudevan, Q.~V. Le, and H.~Adam.
\newblock Searching for mobilenetv3.
\newblock {\em CoRR}, abs/1905.02244, 2019.

\bibitem{Weyand17_mobilenet}
A.~G. Howard, M.~Zhu, B.~Chen, D.~Kalenichenko, W.~Wang, T.~Weyand,
  M.~Andreetto, and H.~Adam.
\newblock Mobilenets: Efficient convolutional neural networks for mobile vision
  applications.
\newblock {\em CoRR}, abs/1704.04861, 2017.

\bibitem{Huang2016SpeedAccuracyTF}
J.~Huang, V.~Rathod, C.~Sun, M.~Zhu, A.~K. Balan, A.~Fathi, I.~Fischer,
  Z.~Wojna, Y.~Song, S.~Guadarrama, and K.~P. Murphy.
\newblock Speed/accuracy trade-offs for modern convolutional object detectors.
\newblock {\em CVPR}, pages 3296--3297, 2017.

\bibitem{Huang17_tf_api}
J.~Huang, V.~Rathod, C.~Sun, M.~Zhu, A.~Korattikara, A.~Fathi, I.~Fischer,
  Z.~Wojna, Y.~Song, S.~Guadarrama, and K.~Murphy.
\newblock Speed/accuracy trade-offs for modern convolutional object detectors.
\newblock In {\em CVPR}, 2017.

\bibitem{yolov3_github}
G.~Jocher.
\newblock {YOLOv3 in PyTorch}.
\newblock online: \url{https://github.com/ultralytics/yolov3}, September 2019.

\bibitem{He16_resnet1k}
S.~R. J.~S. Kaiming~He, Xiangyu~Zhang.
\newblock {Deep Residual Networks with 1K Layers}.
\newblock online: \url{https://github.com/KaimingHe/resnet-1k-layers}, April
  2016.

\bibitem{Kang17_tpn}
K.~Kang, H.~Li, T.~Xiao, W.~Ouyang, J.~Yan, X.~Liu, and X.~Wang.
\newblock Object detection in videos with tubelet proposal networks.
\newblock In {\em CVPR}, pages 889--897. IEEE Computer Society, 2017.

\bibitem{Kang18_tcnn}
K.~Kang, H.~Li, J.~Yan, X.~Zeng, B.~Yang, T.~Xiao, C.~Zhang, Z.~Wang, R.~Wang,
  X.~Wang, and W.~Ouyang.
\newblock T-cnn: Tubelets with convolutional neural networks for object
  detection from videos.
\newblock {\em IEEE Transactions on Circuits and Systems for Video Technology},
  pages 1--1, 2018.

\bibitem{Kang16_tcnn}
K.~Kang, W.~Ouyang, H.~Li, and X.~Wang.
\newblock Object detection from video tubelets with convolutional neural
  networks.
\newblock In {\em CVPR}, pages 817--825, June 2016.

\bibitem{RectLabel}
R.~Kawamura.
\newblock Rectlabel: An image annotation tool to label images for bounding box
  object detection and segmentation.
\newblock online: https://rectlabel.com/, 2019.

\bibitem{Kellenberger19_uav}
B.~Kellenberger, D.~Marcos, S.~Lobry, and D.~Tuia.
\newblock Half a percent of labels is enough: Efficient animal detection in
  {UAV} imagery using deep cnns and active learning.
\newblock {\em CoRR}, abs/1907.07319, 2019.

\bibitem{Kellenberger2018_dataset_uav}
B.~Kellenberger, D.~Marcos, and D.~Tuia.
\newblock Detecting mammals in uav images: Best practices to address a
  substantially imbalanced dataset with deep learning.
\newblock {\em Remote Sensing of Environment}, 216:139 -- 153, 2018.

\bibitem{Kellenberger18_dataset_uav}
B.~Kellenberger, D.~Marcos~Gonzalez, and D.~Tuia.
\newblock Best practices to train deep models on imbalanced datasets - a case
  study on animal detection in aerial imagery.
\newblock In {\em European Conference on Machine Learning and Principles and
  Practice of Knowledge Discovery in Databases}, 01 2018.

\bibitem{Kellenberger17_uav}
B.~{Kellenberger}, M.~{Volpi}, and D.~{Tuia}.
\newblock Fast animal detection in uav images using convolutional neural
  networks.
\newblock In {\em IEEE International Geoscience and Remote Sensing Symposium
  (IGARSS)}, pages 866--869, July 2017.

\bibitem{Khosla_FGVC2011_dataset_dog}
A.~Khosla, N.~Jayadevaprakash, B.~Yao, and L.~Fei-Fei.
\newblock Novel dataset for fine-grained image categorization.
\newblock In {\em First Workshop on Fine-Grained Visual Categorization, CVPR},
  Colorado Springs, CO, June 2011.

\bibitem{Kuznetsova18ax_OpenImages}
A.~Kuznetsova, H.~Rom, N.~Alldrin, J.~Uijlings, I.~Krasin, J.~Pont-Tuset,
  S.~Kamali, S.~Popov, M.~Malloci, T.~Duerig, and V.~Ferrari.
\newblock The open images dataset v4: Unified image classification, object
  detection, and visual relationship detection at scale.
\newblock {\em arXiv:1811.00982}, 2018.

\bibitem{Li19_tiger_dataset}
S.~Li, J.~Li, W.~Lin, and H.~Tang.
\newblock Amur tiger re-identification in the wild.
\newblock {\em CoRR}, abs/1906.05586, 2019.

\bibitem{Li17_fcis}
Y.~Li, H.~Qi, J.~Dai, X.~Ji, and Y.~Wei.
\newblock Fully convolutional instance-aware semantic segmentation.
\newblock In {\em CVPR}, pages 4438--4446, 2017.

\bibitem{Lin2016FeaturePN}
T.-Y. Lin, P.~Doll{\'a}r, R.~B. Girshick, K.~He, B.~Hariharan, and S.~J.
  Belongie.
\newblock Feature pyramid networks for object detection.
\newblock {\em CVPR}, pages 936--944, 2017.

\bibitem{Lin2017FocalLF}
T.-Y. Lin, P.~Goyal, R.~B. Girshick, K.~He, and P.~Doll{\'a}r.
\newblock Focal loss for dense object detection.
\newblock {\em ICCV}, pages 2999--3007, 2017.

\bibitem{TsungYi14_coco}
T.-Y. Lin, M.~Maire, S.~Belongie, J.~Hays, P.~Perona, D.~Ramanan,
  P.~Doll{\'a}r, and C.~L. Zitnick.
\newblock Microsoft coco: Common objects in context.
\newblock In D.~Fleet, T.~Pajdla, B.~Schiele, and T.~Tuytelaars, editors, {\em
  ECCV 2014}, pages 740--755, Cham, 2014. Springer International Publishing.

\bibitem{Liu18_review}
L.~Liu, W.~Ouyang, X.~Wang, P.~W. Fieguth, J.~Chen, X.~Liu, and
  M.~Pietik{\"{a}}inen.
\newblock Deep learning for generic object detection: {A} survey.
\newblock {\em CoRR}, abs/1809.02165, 2018.

\bibitem{Liu18_vid_lstm}
M.~Liu and M.~Zhu.
\newblock Mobile video object detection with temporally-aware feature maps.
\newblock {\em CVPR}, pages 5686--5695, 2018.

\bibitem{Liu19_vid_lstm}
M.~Liu, M.~Zhu, M.~White, Y.~Li, and D.~Kalenichenko.
\newblock Looking fast and slow: Memory-guided mobile video object detection.
\newblock {\em CoRR}, abs/1903.10172, 2019.

\bibitem{Liu16_ssd}
W.~Liu, D.~Anguelov, D.~Erhan, C.~Szegedy, S.~Reed, C.-Y. Fu, and A.~C. Berg.
\newblock Ssd: Single shot multibox detector.
\newblock In B.~Leibe, J.~Matas, N.~Sebe, and M.~Welling, editors, {\em
  Computer Vision -- ECCV 2016}, pages 21--37, Cham, 2016. Springer
  International Publishing.

\bibitem{Mnck2018BioTrackerAO}
H.~J. M{\"o}nck, A.~J{\"o}rg, T.~von Falkenhausen, J.~Tanke, B.~Wild,
  D.~Dormagen, J.~Piotrowski, C.~Winklmayr, D.~Bierbach, and T.~Landgraf.
\newblock Biotracker: An open-source computer vision framework for visual
  animal tracking.
\newblock {\em ArXiv}, abs/1803.07985, 2018.

\bibitem{Norouzzadeh2018_trap}
M.~S. Norouzzadeh, A.~Nguyen, M.~Kosmala, A.~Swanson, M.~S. Palmer, C.~Packer,
  and J.~Clune.
\newblock Automatically identifying, counting, and describing wild animals in
  camera-trap images with deep learning.
\newblock {\em Proceedings of the National Academy of Sciences},
  115(25):E5716--E5725, 2018.

\bibitem{Pan2010ASO_transfer}
S.~J. Pan and Q.~Yang.
\newblock A survey on transfer learning.
\newblock {\em IEEE Transactions on Knowledge and Data Engineering},
  22:1345--1359, 2010.

\bibitem{Parham18_wild_dataset}
J.~{Parham}, C.~{Stewart}, J.~{Crall}, D.~{Rubenstein}, J.~{Holmberg}, and
  T.~{Berger-Wolf}.
\newblock An animal detection pipeline for identification.
\newblock In {\em WACV}, 2018.

\bibitem{parkhi12a_iiit_pet_dataset}
O.~M. Parkhi, A.~Vedaldi, A.~Zisserman, and C.~V. Jawahar.
\newblock Cats and dogs.
\newblock In {\em CVPR}, 2012.

\bibitem{Patman18_BioSense}
J.~{Patman}, S.~C.~J. {Michael}, M.~M.~F. {Lutnesky}, and K.~{Palaniappan}.
\newblock Biosense: Real-time object tracking for animal movement and behavior
  research.
\newblock In {\em 2018 IEEE Applied Imagery Pattern Recognition Workshop
  (AIPR)}, pages 1--8, Oct 2018.

\bibitem{Pinheiro15_deepmask}
P.~H.~O. Pinheiro, R.~Collobert, and P.~Doll{\'{a}}r.
\newblock Learning to segment object candidates.
\newblock In {\em NIPS}, 2015.

\bibitem{Pinheiro16_sharpmask}
P.~O. Pinheiro, T.~Lin, R.~Collobert, and P.~Doll{\'{a}}r.
\newblock Learning to refine object segments.
\newblock In {\em ECCV}, 2016.

\bibitem{Redmon_darknet}
J.~Redmon.
\newblock Darknet.
\newblock online: \url{https://github.com/pjreddie/darknet}, August 2018.

\bibitem{Redmon15_yolo}
J.~Redmon, S.~K. Divvala, R.~B. Girshick, and A.~Farhadi.
\newblock {You Only Look Once: Unified, Real-Time Object Detection}.
\newblock {\em CVPR}, pages 779--788, 2015.

\bibitem{Redmon2016_yolov2}
J.~Redmon and A.~Farhadi.
\newblock {YOLO9000: Better, Faster, Stronger}.
\newblock {\em CVPR}, pages 6517--6525, 2017.

\bibitem{Redmon18_yolov3}
J.~Redmon and A.~Farhadi.
\newblock {YOLOv3: An Incremental Improvement}.
\newblock {\em CoRR}, abs/1804.02767, 2018.

\bibitem{Shaoqing15_frcnn}
S.~Ren, K.~He, R.~Girshick, and J.~Sun.
\newblock Faster r-cnn: Towards real-time object detection with region proposal
  networks.
\newblock In {\em Proceedings of the 28th International Conference on Neural
  Information Processing Systems - Volume 1}, NIPS'15, pages 91--99, Cambridge,
  MA, USA, 2015. MIT Press.

\bibitem{Ren17_frcnn}
S.~{Ren}, K.~{He}, R.~{Girshick}, and J.~{Sun}.
\newblock {Faster R-CNN: Towards Real-Time Object Detection with Region
  Proposal Networks}.
\newblock {\em IEEE Transactions on Pattern Analysis and Machine Intelligence},
  39(6):1137--1149, June 2017.

\bibitem{Rey17_uav}
N.~Rey, M.~Volpi, S.~Joost, and D.~Tuia.
\newblock Detecting animals in african savanna with uavs and the crowds.
\newblock {\em ArXiv}, abs/1709.01722, 2017.

\bibitem{Ferrero19_vid_idtracker}
F.~Romero-Ferrero, M.~G. Bergomi, R.~C. Hinz, F.~J.~H. Heras, and G.~G.
  de~Polavieja.
\newblock idtracker.ai: tracking all individuals in small or large collectives
  of unmarked animals.
\newblock {\em Nature Methods}, 16(2):179--182, 2019.

\bibitem{Ronneberger15_unet}
O.~Ronneberger, P.~Fischer, and T.~Brox.
\newblock U-net: Convolutional networks for biomedical image segmentation.
\newblock In {\em {MICCAI} {(3)}}, volume 9351 of {\em Lecture Notes in
  Computer Science}, pages 234--241. Springer, 2015.

\bibitem{Russell19_LabelMe}
B.~Russell, A.~Torralba, J.~Yuen, and et. al.
\newblock Labelme: The open annotation tool.
\newblock online: http://labelme2.csail.mit.edu/Release3.0/index.php, 2019.

\bibitem{Sandler18_mobilenetv2}
M.~Sandler, A.~G. Howard, M.~Zhu, A.~Zhmoginov, and L.~Chen.
\newblock Mobilenetv2: Inverted residuals and linear bottlenecks.
\newblock In {\em CVPR}, pages 4510--4520, 2018.

\bibitem{Schneider18_trap}
S.~Schneider, G.~W. Taylor, and S.~C. Kremer.
\newblock Deep learning object detection methods for ecological camera trap
  data.
\newblock {\em CoRR}, abs/1803.10842, 2018.

\bibitem{Schneider19_trap_review}
S.~Schneider, G.~W. Taylor, S.~Linquist, and S.~C. Kremer.
\newblock Past, present and future approaches using computer vision for animal
  reidentification from camera trap data.
\newblock {\em Methods in Ecology and Evolution}, 10(4):461--470, April 2018.

\bibitem{Sridhar19_vid_Tracktor}
V.~H. Sridhar, D.~G. Roche, and S.~Gingins.
\newblock Tracktor: Image-based automated tracking of animal movement and
  behaviour.
\newblock {\em Methods in Ecology and Evolution}, 10(6):815--820, 2019.

\bibitem{Swanson2015_serengeti_dataset}
A.~Swanson, M.~Kosmala, C.~Lintott, R.~Simpson, A.~Smith, and C.~Packer.
\newblock Snapshot serengeti, high-frequency annotated camera trap images of 40
  mammalian species in an african savanna.
\newblock {\em Scientific Data}, 2:150026--, June 2015.

\bibitem{Szegedy17_inceptionv4}
C.~Szegedy, S.~Ioffe, V.~Vanhoucke, and A.~A. Alemi.
\newblock Inception-v4, inception-resnet and the impact of residual connections
  on learning.
\newblock In {\em Proceedings of the Thirty-First {AAAI} Conference on
  Artificial Intelligence, February 4-9, 2017, San Francisco, California,
  {USA.}}, pages 4278--4284, 2017.

\bibitem{Szegedy15_inception}
C.~Szegedy, W.~Liu, Y.~Jia, P.~Sermanet, S.~E. Reed, D.~Anguelov, D.~Erhan,
  V.~Vanhoucke, and A.~Rabinovich.
\newblock Going deeper with convolutions.
\newblock In {\em CVPR}, 2015.

\bibitem{Szegedy2013DeepNN}
C.~Szegedy, A.~Toshev, and D.~Erhan.
\newblock Deep neural networks for object detection.
\newblock In {\em NIPS}, 2013.

\bibitem{Szegedy16_inceptionv2}
C.~Szegedy, V.~Vanhoucke, S.~Ioffe, J.~Shlens, and Z.~Wojna.
\newblock Rethinking the inception architecture for computer vision.
\newblock In {\em CVPR}, pages 2818--2826, 2016.

\bibitem{Tabak18_camera_trap_north_america}
M.~A. Tabak, Mohammad, S.~Norouzzadeh, and D.~W.~W. et. al.
\newblock Machine learning to classify animal species in camera trap images:
  Applications in ecology.
\newblock {\em Methods in Ecology and Evolution}, 10(4):585--590, 26 November
  2018.

\bibitem{TerSarkisov2018_vid_beef_cattle}
A.~Ter-Sarkisov, J.~D. Kelleher, B.~Earley, M.~Keane, and R.~J. Ross.
\newblock Beef cattle instance segmentation using fully convolutional neural
  network.
\newblock In {\em BMVC}, 2018.

\bibitem{darkflow_github}
Trieu.
\newblock {Darkflow}.
\newblock online: \url{https://github.com/thtrieu/darkflow}, March 2018.

\bibitem{active_learning11}
D.~{Tuia}, M.~{Volpi}, L.~{Copa}, M.~{Kanevski}, and J.~{Munoz-Mari}.
\newblock A survey of active learning algorithms for supervised remote sensing
  image classification.
\newblock {\em IEEE Journal of Selected Topics in Signal Processing},
  5(3):606--617, June 2011.

\bibitem{Tzutalin15_LabelImg}
Tzutalin.
\newblock Labelimg: A graphical image annotation tool and label object bounding
  boxes in images.
\newblock online: https://github.com/tzutalin/labelImg, 2015.

\bibitem{labelme2016}
K.~Wada.
\newblock {labelme: Image Polygonal Annotation with Python}.
\newblock \url{https://github.com/wkentaro/labelme}, 2016.

\bibitem{DaSiamRPN_github}
Q.~Wang.
\newblock {DaSiamRPN}.
\newblock online: \url{https://github.com/foolwood/DaSiamRPN}, September 2019.

\bibitem{SiamMask_github}
Q.~Wang.
\newblock {SiamMask}.
\newblock online: \url{https://github.com/foolwood/SiamMask}, September 2019.

\bibitem{wang2019_siam_mask}
Q.~Wang, L.~Zhang, L.~Bertinetto, W.~Hu, and P.~H. Torr.
\newblock Fast online object tracking and segmentation: A unifying approach.
\newblock In {\em CVPR}, 2019.

\bibitem{Wang16_class_imbalance}
S.~{Wang}, W.~{Liu}, J.~{Wu}, L.~{Cao}, Q.~{Meng}, and P.~J. {Kennedy}.
\newblock Training deep neural networks on imbalanced data sets.
\newblock In {\em International Joint Conference on Neural Networks}, 2016.

\bibitem{WelinderEtal2010_caltech_usd_birds}
P.~Welinder, S.~Branson, T.~Mita, C.~Wah, F.~Schroff, S.~Belongie, and
  P.~Perona.
\newblock {Caltech-UCSD Birds 200}.
\newblock Technical Report CNS-TR-2010-001, California Institute of Technology,
  2010.

\bibitem{Werkhoven19_vid_margo}
Z.~Werkhoven, C.~Rohrsen, C.~Qin, B.~Brembs, and B.~de~Bivort.
\newblock Margo (massively automated real-time gui for object-tracking), a
  platform for high-throughput ethology.
\newblock {\em bioRxiv}, 2019.

\bibitem{Wilkins2019_avc}
D.~C. Wilkins, K.~M. Kockelman, and N.~Jiang.
\newblock Animal-vehicle collisions in texas: How to protect travelers and
  animals on roadways.
\newblock {\em Accident Analysis \& Prevention}, 131:157--170, 2019.

\bibitem{Willi18_camera_trap_data}
M.~Willi, R.~T. Pitman, and A.~W.~C. et. al.
\newblock Identifying animal species in camera trap images using deep learning
  and citizen science.
\newblock {\em Methods in Ecology and Evolution}, 2018.

\bibitem{Xie15_hed}
S.~{Xie} and Z.~{Tu}.
\newblock Holistically-nested edge detection.
\newblock In {\em ICCV}, pages 1395--1403, Dec 2015.

\bibitem{Xue17_avc_wireless}
W.~{Xue}, T.~{Jiang}, and J.~{Shi}.
\newblock Animal intrusion detection based on convolutional neural network.
\newblock In {\em 2017 17th International Symposium on Communications and
  Information Technologies (ISCIT)}, pages 1--5, Sep. 2017.

\bibitem{semantic_cvpr19}
F.~A. R. K. J. S. S. N. A. T. B.~C. Yi~Zhu*, Karan~Sapra*.
\newblock Improving semantic segmentation via video propagation and label
  relaxation.
\newblock In {\em CVPR}, June 2019.

\bibitem{Yosinski14_nips_transfer}
J.~Yosinski, J.~Clune, Y.~Bengio, and H.~Lipson.
\newblock How transferable are features in deep neural networks?
\newblock In {\em NIPS}. 2014.

\bibitem{Yousif17_trap}
H.~{Yousif}, J.~{Yuan}, R.~{Kays}, and Z.~{He}.
\newblock Fast human-animal detection from highly cluttered camera-trap images
  using joint background modeling and deep learning classification.
\newblock In {\em 2017 IEEE International Symposium on Circuits and Systems
  (ISCAS)}, pages 1--4, May 2017.

\bibitem{Hayder18_trap}
H.~Yousif, J.~Yuan, R.~Kays, and Z.~He.
\newblock Object detection from dynamic scene using joint background modeling
  and fast deep learning classification.
\newblock {\em Journal of Visual Communication and Image Representation},
  55:802--815, 2018.

\bibitem{zhou2017_ade20k}
B.~Zhou, H.~Zhao, X.~Puig, S.~Fidler, A.~Barriuso, and A.~Torralba.
\newblock Scene parsing through ade20k dataset.
\newblock In {\em CVPR}, 2017.

\bibitem{Zhu17_trap}
C.~{Zhu}, T.~H. {Li}, and G.~{Li}.
\newblock Towards automatic wild animal detection in low quality camera-trap
  images using two-channeled perceiving residual pyramid networks.
\newblock In {\em ICCVW}, pages 2860--2864, Oct 2017.

\bibitem{Zhu15_kinect}
Q.~{Zhu}, J.~{Ren}, D.~{Barclay}, S.~{McCormack}, and W.~{Thomson}.
\newblock Automatic animal detection from kinect sensed images for livestock
  monitoring and assessment.
\newblock In {\em IEEE International Conference on Computer and Information
  Technology}, pages 1154--1157, Oct 2015.

\bibitem{Zhu17_fgfa}
X.~Zhu, Y.~Wang, J.~Dai, L.~Yuan, and Y.~Wei.
\newblock Flow-guided feature aggregation for video object detection.
\newblock In {\em ICCV}, pages 408--417, Oct 2017.

\bibitem{Zhu17_deep_feature_flow}
X.~Zhu, Y.~Xiong, J.~Dai, L.~Yuan, and Y.~Wei.
\newblock Deep feature flow for video recognition.
\newblock In {\em CVPR}, pages 4141--4150, July 2017.

\bibitem{Zhu18_dasiamrpn}
Z.~Zhu, Q.~Wang, L.~Bo, W.~Wu, J.~Yan, and W.~Hu.
\newblock {Distractor-aware Siamese Networks for Visual Object Tracking}.
\newblock In {\em ECCV}, 2018.

\bibitem{Zin16_vid_cow}
T.~T. {Zin}, I.~{Kobayashi}, P.~{Tin}, and H.~{Hama}.
\newblock A general video surveillance framework for animal behavior analysis.
\newblock In {\em 2016 Third International Conference on Computing Measurement
  Control and Sensor Network (CMCSN)}, pages 130--133, May 2016.

\bibitem{Zoph17_nas_rl}
B.~Zoph and Q.~V. Le.
\newblock Neural architecture search with reinforcement learning.
\newblock In {\em ICLR}, 2017.

\bibitem{Zoph18_nas}
B.~Zoph, V.~Vasudevan, J.~Shlens, and Q.~V. Le.
\newblock Learning transferable architectures for scalable image recognition.
\newblock In {\em CVPR}, pages 8697--8710, 2018.

\end{thebibliography}
}

\end{document}